\documentclass{article}

\usepackage[final]{nips_2018}

\usepackage[utf8]{inputenc} % allow utf-8 input
\usepackage[T1]{fontenc}    % use 8-bit T1 fonts
\usepackage{hyperref}       % hyperlinks
\usepackage{url}            % simple URL typesetting
\usepackage{booktabs}       % professional-quality tables
\usepackage{amsfonts}       % blackboard math symbols
\usepackage{nicefrac}       % compact symbols for 1/2, etc.
\usepackage{microtype}      % microtypography

\usepackage{endnotes}
\usepackage{amsmath}
\usepackage[utf8]{inputenc}
\usepackage[T1]{fontenc}
\usepackage[english]{babel}
\usepackage{color}

\newcommand{\ignore}[1]{}
\newcommand{\fixlater}[1]{}

\newcommand{\eat}[1]{}

\usepackage{subfig}
\usepackage{graphicx}
\usepackage{multirow}
\usepackage{endnotes}
\usepackage{epsfig}
\usepackage{mdframed}
\usepackage{amsmath}
\usepackage{amsfonts}
\usepackage{amssymb}
\usepackage{mathrsfs}
\usepackage{theorem}
\usepackage{mathtools}

\usepackage{pifont}

\usepackage{xspace}
\usepackage{listings}
\usepackage{lastpage}

\usepackage{lipsum,caption,graphicx}
\usepackage{float}
\usepackage{subfig}

\usepackage{tikz}
\usetikzlibrary{er,positioning}
\usetikzlibrary{trees}

\begin{document}

\title{The Big Three: A Methodology to Increase Data Science ROI by Answering the Questions Companies Care About \\
}

% \author{\IEEEauthorblockN{Daniel K. Griffin}
% \IEEEauthorblockA{Cisco Senior Data Scientist \\
% \textit{dagriff2@cisco.com}}
% % \and
% % \IEEEauthorblockN{2\textsuperscript{nd} Given Name Surname}
% % \IEEEauthorblockA{\textit{dept. name of organization (of Aff.)} \\
% % \textit{name of organization (of Aff.)}\\
% % City, Country \\
% % email address}
% % \and
% % \IEEEauthorblockN{3\textsuperscript{rd} Given Name Surname}
% % \IEEEauthorblockA{\textit{dept. name of organization (of Aff.)} \\
% % \textit{name of organization (of Aff.)}\\
% % City, Country \\
% % email address}
% % \and
% % \IEEEauthorblockN{4\textsuperscript{th} Given Name Surname}
% % \IEEEauthorblockA{\textit{dept. name of organization (of Aff.)} \\
% % \textit{name of organization (of Aff.)}\\
% % City, Country \\
% % email address}
% % \and
% % \IEEEauthorblockN{5\textsuperscript{th} Given Name Surname}
% % \IEEEauthorblockA{\textit{dept. name of organization (of Aff.)} \\
% % \textit{name of organization (of Aff.)}\\
% % City, Country \\
% % email address}
% % \and
% % \IEEEauthorblockN{6\textsuperscript{th} Given Name Surname}
% % \IEEEauthorblockA{\textit{dept. name of organization (of Aff.)} \\
% % \textit{name of organization (of Aff.)}\\
% % City, Country \\
% % email address}
% }

\author{%
  Daniel K. Griffin \\
  Staff Data Scientist \\
  \texttt{dagriff2@cisco.com}
  % examples of more authors
  % Coauthor \\
  % Affiliation \\
  % Address \\
  % \texttt{email} \\
  % \AND
  % Coauthor \\
  % Affiliation \\
  % Address \\
  % \texttt{email} \\
  % \And
  % Coauthor \\
  % Affiliation \\
  % Address \\
  % \texttt{email} \\
  % \And
  % Coauthor \\
  % Affiliation \\
  % Address \\
  % \texttt{email} \\
}

\maketitle

\begin{abstract}

    Companies may be achieving only a third of the value they could be getting from data science in industry applications. In this paper, we propose a methodology for categorizing and answering 'The Big Three' questions (what is going on, what is causing it, and what actions can I take that will optimize what I care about) using data science. The applications of data science seem to be nearly endless in today's modern landscape, with each company jockeying for position in the new data and insights economy. Yet, data scientists seem to be solely focused on using classification, regression, and clustering methods to answer the question 'what is going on'. Answering questions about why things are happening or how to take optimal actions to improve metrics are relegated to niche fields of research and generally neglected in industry data science analysis. We survey technical methods to answer these other important questions, describe areas in which some of these methods are being applied, and provide a practical example of how to apply our methodology and selected methods to a real business use case.
    
\end{abstract}

\section{Introduction}

    In the past decade, there has been an explosion in the application of data science outside of academic realms. The use of general, statistical, predictive machine learning models has achieved high success rates across multiple occupations including finance, marketing, sales, and engineering, as well as multiple industries including entertainment, online and store front retail, transportation, service and hospitality, healthcare, insurance, manufacturing and many others. The applications of data science seem to be nearly endless in today's modern landscape, with each company jockeying for position in the new data and insights economy. Yet, what if I told you that companies may be achieving only a third of the value they could be getting with the use of data science for their companies? I know, it sounds almost fantastical given how much success has already been achieved using data science. However, many opportunities for value generation may be getting overlooked because data scientists and statisticians are not traditionally trained to answer some of the questions companies in industry care about. 
    
    Most of the technical data science analysis done today is either classification (labeling with discrete values), regression (labeling with a number), or pattern recognition. These forms of analysis answer the business questions 'can I understand what is going on' and 'can I predict what will happen next'. Examples of questions are 'can I predict which customers will renew a service contract?', 'can I forecast my next quarter revenue?', 'can I predict products customers are interested in?', 'are there important customer activity patterns?', etc. These are extremely valuable questions companies care about that can be answered using data science. In fact, answering these questions is what has caused the explosion of interest in applying data science in business applications. However, most companies have two other major categories of important questions that are being largely ignored. Namely, once a problem has been identified or predicted, can we determine what's causing it? Furthermore, once the cause of an issue has been determined, can we take action to resolve or prevent the problem?
    
    This paper starts by outlining what 'The Big Three' questions companies in industry applications care about are in section 2. Section 3 provides an overview of how the methods for answering the big three questions are categorized. Section 4 is an outline for sections 5, 6, and 7. Section 5 discusses classic machine learning methods for answering 'what is happening'. Section 6 discusses causal analysis (we use the terms causal analysis and causal inference interchangeably) for answering 'why is something happening'. Section 7 discusses methods for learning how to act intelligently for answering 'what actions should be taken to improve what I care about'.
    Sections 5, 6, and 7 each have two subsections. The first subsection details specific methods and algorithms that can be used to answer the question of interest. The second subsection references some example applications of answering the question of interest. Section 8 takes the real world business problem of predicting whether or not a customer will renew a service contract that they have with a company. Section 9 concludes the paper with a recap of all the major points described in the paper.

\section{The Big Three}

    \begin{figure}[h]
      \centering
        \includegraphics[width=1\textwidth]{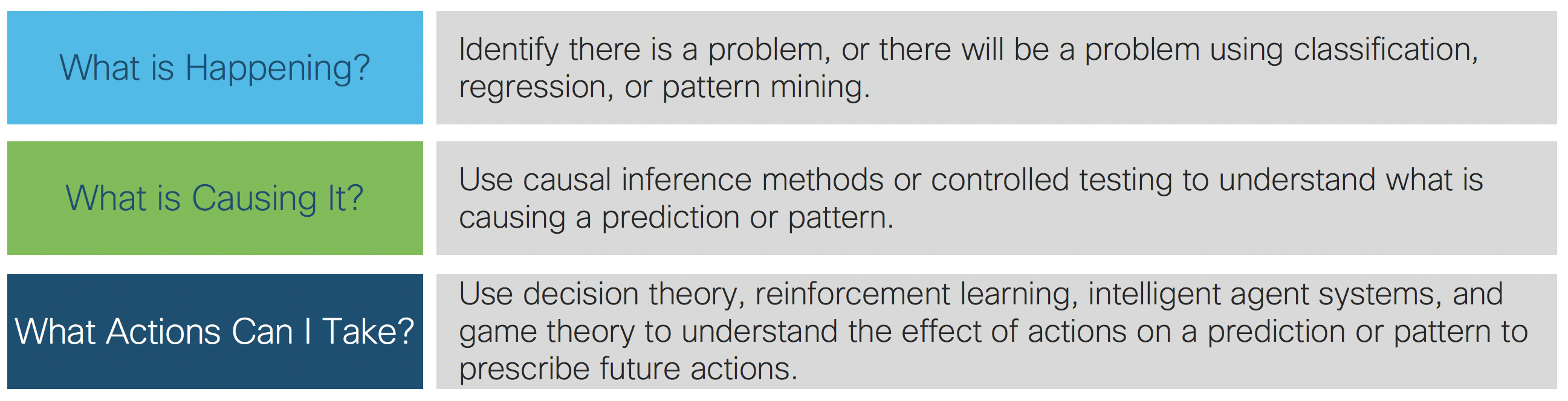}
      \caption{The Big Three Questions}
    \end{figure}

    Figure 1 above describes what 'The Big Three' questions are. The big three questions seem fairly obvious. In fact, these questions are at the foundation of most of problem solving. Yet, almost all data science in industry today revolves around answering only the first question. What most data scientists understand as supervised, unsupervised, and semi-supervised learning revolves around answering what is happening or what will happen. Even with something like a product recommendation system (which you might believe prescribes something because of the term 'recommend'), we only get what products a customer is predicted to be interested in (thus it is a prediction or pattern mining method). We don't know the most effective way to act on that predicted information. Should we send an ad? Should we call them? Do certain engagements with them decrease their chances of purchase? To answer what is causing something to happen, we need to rely on methods of controlled testing and causal inference. Once we understand what is causing a metric we care about, we can begin to think intelligently about the actions we can take to change those metrics. This is where the third question mentioned in figure 1 above comes in. To answer this question we can rely on a wide variety of techniques that have been developed including causal inference for the cause and effect relationship between actions and the metrics they are supposed to affect, decision theory, reinforcement learning, intelligent agents, and game theory.
    
    It's also important to understand that there are two contexts under which we can consider the big three questions. The first context is the reactionary context. In the reactionary context, the questions are 'what is happening', 'why is it happening', and 'what actions can I take to halt or encourage it'. These questions assume that the analysis being performed is to understand the current state of some system and to act in a reactionary fashion. The second context is the proactive context. In the proactive context, the questions are 'what will happen', 'why will it happen', and 'what actions can I take to prevent or encourage it'. These questions assume that the analysis being performed is to understand the future state of some system and to act in a proactive fashion. Regardless of the context, the three questions remain the same in spirit. In the next section, we show a categorization of methods and fields related to answering each of the big three questions.

\section{Categorization of Methods}

    \begin{table*}[ht]
        \begin{center}
            \begin{tabular}{c c c}
             \hline
             Prediction and Pattern Mining & Causality & Intelligent Actions \\
             \hline
             Classification & A/B Testing & Decision Theory \\
             Regression & Difference in Differences & Influence Diagrams \\
             Clustering & Regression Discontinuity Design & Reinforcement Learning  \\
             Outlier Detection & Uplift Modeling & Game Theory and Multiple Agents \\
             Recommendation Systems & Potential Outcomes & \\
             Frequent Pattern Mining & Causal Graphical Models & \\
             Temporal Pattern Mining &  & \\
             \hline
            \end{tabular}
        \end{center}
        \caption{Sampling of Methods and Fields for Answering The Big Three Questions}
    \end{table*}

    In table 1, we provide a brief sampling of methods and fields associated with answering each of the big three questions. These lists are not meant to be totally exhaustive, but rather to provide some general insight into what kinds of analysis loosely fall into answering a particular question. These lists are also not mutually exclusive. For example, regression methods are used a lot in modern approximate methods for reinforcement learning \cite{Sutton1998}, and online regression methods can be framed algorithmically in the same way as reinforcement learning \cite{SL_as_RL}. Causality is also used very often to model the effect of actions taken on the state of a system.
    
    The column titled 'Prediction and Pattern Mining' lists many high level analysis categories that are typically employed to answer the first of the big three questions. These are methods that most data scientists are familiar with, and are outlined in many great resources \cite{Aggarwal:2015:DMT:2778285} \cite{hastie01statisticallearning} \cite{Mitchell:1997:ML:541177} \cite{Aggarwal:2016:RST:2931100}. In section 5 we outline the methodology of supervised and unsupervised learning at a high level. We leave a more in depth exploration of the methods to the other resources we've referenced since these methods are already widely understood and applied in business data science applications.
    
    The column titled 'Causality' lists a few methods and approaches to answering questions related to causality and causal inference. Many of the methods at the top of the list deal with controlled experiments and quasi-experiments. The methods listed at the end deal more with the observational setting where many potential confounding variables exist.
    
    The column titled 'Intelligent Actions' lists many fields (which is a bit different from the other two columns which list categories of analysis, and specific analysis methods) related to understanding how actions impact an environment, and how to optimize actions and decision making in an environment. One could also claim that causal inference methods can also be used to understand the cause and effect relationship between actions and variables of interest, and we would agree. However, we've decided to dedicate a separate column to  causal inference methods. This is to help make the separation of concerns a bit clearer. Most of the methods listed under the Intelligent Actions column implicitly rely on causal relationships. For example, much of the exploration performed in reinforcement learning methods can be thought of as advanced, intelligent controlled testing (akin to A/B testing). To simplify things, we frame causality as answering the question of how one random variable effects another when the random variables do not directly involve actions. For example, let's suppose we are in a hypothetical environment where we have actions which are a choice of food to eat, a random variable which represents carbohydrate intake, and another random variable representing blood glucose levels. We can use causal inference to understand the cause and effect relationship between carbohydrate intake and blood glucose levels. This is the kind of analysis we are thinking of when we specify the Causality column in our table. We could also ask how does taking the action of eating a particular food effect carbohydrate intake. This too is a way to use causal inference to understand the effect of actions, which is typically used in domains like adverse drug reaction discovery. Solving this kind of causal question overlaps with both the Causality and Intelligent Actions sections. The Intelligent Actions column outlines a variety of other methods where causality is implicitly utilized to handle the challenges that arise specifically due to acting in an environment. We maintain this separation so that we can more easily align the methods to the questions companies care about answering.

\section{Methods Outline}
    
    The following is an outline for the methodology sections.
    
    \begin{itemize}
        \item Prediction and Pattern Mining:
            \begin{itemize}
                \item Methods:
                    \begin{itemize}
                        \item Unsupervised Learning
                        \item Supervised Learning
                    \end{itemize}
                \item Applications:
                    \begin{itemize}
                        \item Recommendation Systems
                        \item Sales and Demand Forecasting
                        \item Stock Market Analysis
                    \end{itemize}
        \end{itemize}
        \item Causality:
            \begin{itemize}
                \item Methods:
                    \begin{itemize}
                        \item Controlled Experiments
                        \item Quasi-Experiments
                        \item Observational Studies
                    \end{itemize}
                \item Applications:
                    \begin{itemize}
                        \item Social Sciences and Medicine
                        \item Customer Experience
                    \end{itemize}
            \end{itemize}
        \item Intelligent Action:
            \begin{itemize}
                \item Methods:
                    \begin{itemize}
                        \item Decision Theory
                        \item Influence Diagrams
                        \item Reinforcement Learning
                        \item Game Theory and Multiple Agents
                    \end{itemize}
                \item Applications:
                    \begin{itemize}
                        \item Online and Marketing Applications
                        \item Customer Experience and Engagement
                    \end{itemize}
            \end{itemize}
    \end{itemize}
    
\section{Prediction and Pattern Mining}    

    This section outlines the major methods used to answer questions about prediction and pattern mining. Answering questions about prediction and pattern mining has been widely explored and applied in business. Data science and analysis methods for answering these questions has gained the most ground, and garnered the most success compared to methods for answering the other two questions. There are likely a wide variety of reasons for this including but not limited to: (i) a lack of statistical expertise and experience in the methods required to answer questions about causality, (ii) the lack of deep exposure to reinforcement learning and decision theory, and (iii) a standardization of machine learning and applied statistics curriculums presenting only the classic areas of supervised and unsupervised learning. Thus, we restrict ourselves to exploring unsupervised learning and supervised learning in a broad sense, and leave an in-depth exploration of these topics to the many great resources already available in the field. In this section we also mention 3 major business applications of these data science methods that have been successful. For a much more complete and thorough evaluation of classic AI, machine learning, and data science in practical applications, see this survey done by AAAI \cite{2016_100yr_AI_survey}.

\subsection{Unsupervised Learning}

    The book 'Elements of Statistical Learning' defines unsupervised learning as methods for inferring properties of the joint statistical distribution of a set of data points \cite{hastie01statisticallearning}. Specifically, given a set of \(N\) observations \(\vec{x_1}, ..., \vec{x_N}\) of a random vector \(\vec{X}\), identify characteristics of the joint probability distribution \(Pr(\vec{X})\).
    
    The essential goal of unsupervised learning is the identification and extraction of mathematical patterns from data. Performing this kind of analysis on data typically requires 4 things. The first is the data to analyze. The second is a 'model' of the relationships between data. The third is a 'measure' of how good or bad our model of the relationships in the data set are. The fourth is an optimization method that sets the parameters of our model such that they minimize or maximize the measure of the relationships in data. There isn't much else to say about the data to analyze other than it is typically assumed to be randomly sampled in an independent and identically distributed way (but in some cases like temporal data the data is assumed to be correlated along the time axis in some way). This doesn't have to be the case, but in many cases it is true, and it simplifies the statistical analysis methods. 
    
    The second component, the 'model' of relationships, has received perhaps hundreds of years of treatment across applied mathematics, and can even be thought of as a major pillar of mathematics, science, and the analysis of data. The idea that there can be structured relationships between data values lies at the foundation of every mathematical operator. From the simplest relational forms such as 1+1=2 to the basic statistical regression model of \(Y = mX + b\), math is generally used in the world today to model relationships between data points. A key component of prediction and pattern mining methods is that we assume a model has some parameters that are unknown, and other parameters (eg the data) that are variable. Much of applied mathematics studies the set of choices that can be made (aka equations) that define how data can be related. The choice of model is commonly referred to as setting the 'inductive bias' for the relationships between data points in a data set. Given the infinite set of ways in which we can create models of the relationships between data points, we have to choose a specific set of mathematical equations to model these relationships. Some examples of models of relationships between data points include association rules, clusters, factorized matrices, graphs, and graphical statistical models just to name a few.
    
    The third component of unsupervised learning is the 'measure' or metric for how good or bad our model of the relationships in the data set are. The measure is either framed as a measure of error or a measure of correctness given a models set of parameters. For example, in k-means clustering, the measure of the quality of the relationships described by the model are given as the sum of squared distances (errors) between every data point and it's nearest cluster center. Specifically, 
    
    \begin{center} 
        \begin{equation} \label{eq1}
            error = \frac{1}{2}\sum_{k=1}^{K}\sum_{x_i \in C_k} \left \| x_i - x_{C_k} \right \|
        \end{equation}
    \end{center}

    Where \(k \in 1, ..., K\) is the number of clusters, \(C_k\) represents cluster \(k\), and \(x_{C_k}\) is the vector representing cluster center for cluster \(C_k\). This measure explicitly defines what constitutes a pattern given the k-means clustering model of the relationships in data. For k-means, a pattern is defined as the position of each cluster center, where each cluster center is the representative for all data points nearest to that particular cluster center. The measure of relationships describes how much one one set of cluster center positions describes the data. Using the measure, we can use optimization methods to change the parameters of our model so that the error decreases or the correctness increases.
    
    The fourth and final component of unsupervised learning is the optimization method. The optimization method feeds the data to a model with an initial set of parameter values, and then uses the measure of how well the data relationships are modeled to calculate a concrete numeric representation of the models fit. Using this, the optimization method then suggests a new set of model parameters such that the measure of the model of the data relationships increases or decreases (Depending on how the problem is formulated). This is nothing more than a re-statement of how the field of optimization typically works. All optimization methods are effectively search in one form or another, where the optimization method searches for a configuration of values that minimizes or maximizes some objective. Many different methods for optimization exist including line search, trust region, conjugate gradient, quasi-newton, unconstrained, derivative free, etc. The seminal book 'Numerical Optimization' by Stephen Wright is a great guide to the field of optimization \cite{nocedal_wright_2006}. Most modern machine learning methods use some form of gradient based optimization, where a gradient of the measure of the models fit is taken with respect to the parameters of the model, and this gradient is used to iteratively update model parameters.
    
    We will not expand further on the topic of unsupervised learning in the same way we do for causality and intelligent action in later sections. This field is well studied, and commonly applied to industry and business applications. For the interested reader, we suggest the following resources on unsupervised learning \cite{Russell:2009:AIM:1671238} \cite{hastie01statisticallearning} \cite{Murphy:2012:MLP:2380985} \cite{Mitchell:1997:ML:541177} \cite{Bishop:2006:PRM:1162264} \cite{Aggarwal:2015:DMT:2778285}.

\subsection{Supervised Learning}

    Supervised learning is very similar to unsupervised learning. In fact, the entire framing of the problem is nearly the same. The data, model, measure, and optimization method all still need to be defined. The main difference in the two methods is that supervised learning is used when the goal of modeling is to model the relationship between one or more input variables (covariates, independent variables, etc.), and one or more target variables (outcome, dependent variables, etc.). Specifically, given a set of \(N\) input observations \(\vec{x_1}, ..., \vec{x_N}\) and a set of \(N\) target variables \(\vec{y_1}, ..., \vec{y_N}\) of the random vectors \(X\) and \(Y\), the goal is to identify characteristics of either the joint probability distribution \(Pr(X, Y)\), or the conditional distribution \(Pr(Y | X)\). By framing the problem this way, there is an assumption that the specific kind of relationships of interest are those between \(X\) and \(Y\). Thus, while similar to unsupervised learning, supervised learning restricts the focus of patterns and analysis to those that link the random vectors \(X\) and \(Y\).
    
    Supervised methods have been long studied in the area of applied mathematics. Statistics and statistical inference for regression models is one early field that aimed to model the kinds of relationships found in supervised learning. For example, in linear regression, a set of data pairs \( \left \{ (x_{i1}, ..., x_{im}, y_i) : i \in 1, ..., N \right \} \) is modeled by the equation,
    
    \begin{center}
        \begin{equation} \label{eq2}
            \widehat{y}_i = \beta_0 + \sum_{j=1}^{m}x_{ij}\beta_j 
        \end{equation}
    \end{center}
    
    The coefficients of the model \(\beta_j\) are the parameters of the model. The measure of how good the model is (Residual Sum of Squares),
    
    \begin{center}
        \begin{equation} \label{eq3}
            RSS(\vec{\beta}\ ) = \sum_{i=1}^{N} (y_i - x_{i}^{T}\vec{\beta})^2 
        \end{equation}
    \end{center}
    
    Many optimization methods can be selected to solve this problem, but it turns out that the coefficients \(\beta_j\) can be found analytically. In general, supervised learning problems are framed such that the goal of optimization is to minimize the average error, or maximize the average correctness of the ability of the model to generate the values of random vector \(Y\) given random vector \(X\). Specifically, 
    
    \begin{center}
        \begin{equation} \label{eq4}
            \theta^* = \underset{\theta}{argmin} E_{Y|X} [L(Y,\theta)] 
        \end{equation}
    \end{center}
    
    Where \(\theta\) represents the model parameters, \(\theta^*\) represents the optimal model parameters, and \(E_{Y|X}\) is the expectation of the value in the brackets with respect to the random variable \(Y\) given the random variable \(X\) is fixed. There are many choices for the model such as generalized linear models, support vector machines, adaptive basis function models, graphical statistical models, artificial neural networks, and many others. There are also many choices for the measure of how well the model fits the data including mean squared error, hinge loss, logistic loss, cross-entropy error, and many others.
    
    We will not expand further on the topic of supervised learning in the same way we do for causality and intelligent action in later sections. This field is well studied, and commonly applied to industry and business applications. For the interested reader, we suggest the following resources on supervised learning \cite{Russell:2009:AIM:1671238} \cite{hastie01statisticallearning} \cite{Murphy:2012:MLP:2380985} \cite{Mitchell:1997:ML:541177} \cite{Bishop:2006:PRM:1162264} \cite{Aggarwal:2015:DMT:2778285}.

\subsection{Example Applications}

\subsubsection{Recommendation Systems}

    One area where prediction and pattern mining methods are applied is for development of recommendation systems. The goal of recommendation systems is to use historical preference and behavioral information about users or customers to build a model that can recommend new items that a user might be interested in. For example, the online video streaming company Netflix uses supervised and unsupervised learning methods to build models that can recommend new online video content that a user has not yet watched, but may be interested in. Originally, the problem of recommendation systems can be framed as an unsupervised learning problem (as a matrix factorization problem) where similar customers are grouped together, and the non-overlapping items from the group of similar customers are used as recommendations. As the field has evolved, more complex supervised methods have been developed, and combinations of multiple methods have been introduced. 
    
    Netflix uses a combination of models to recommend new videos to their users \cite{Gomez-Uribe:2015:NRS:2869770.2843948}. The models they use include a personalized video ranker (PVR) which ranks every video in the video catalogue at an individual user level, a Top-N video ranker which ranks the top N videos for a user, a 'trending now' model that builds temporally relevant recommendations based on recent general video activity, a 'continue watching' system that ranks videos that have already been started by a user, a video-video similarity recommendation, a page generation and layout recommendation, and a few others. These algorithms are validated and tested using tuned methods of A/B testing to ensure performance. The system is generally validated from a business perspective by analyzing it's correlation with retention rates. They use an A/B test for measuring the retention change given the change in recommendation algorithm to validate their business impact. In the section on causality, we discuss some challenges with A/B testing and outline alternatives for measuring the impact of actions, decisions, and models on business metrics. 
    Netflix is just one of many companies that are capitalizing on prediction and pattern mining algorithms to improve their company's bottom line. For readers interested in learning more about recommendation systems, the book by Charu Aggarwal is considered by many to be the handbook reference of modern recommendation system algorithms \cite{Aggarwal:2016:RST:2931100}.

\subsubsection{Sales and Demand Forecasting}

    Sales and demand forecasting is another example of how prediction and pattern mining algorithms has achieved large success. The goal of demand forecasting is to predict the total amount of demand that will exist for a particular product, service, or resource. The goal of sales forecasting is to predict the total dollar amount of sales that will occur for a particular product, service, or resource. While similar, these problems are not exactly the same. For example, there may be a predicted demand of 300 million units of network switches for the networking industry in the next quarter. However, a company may be more interested in the amount of sales that a specific company may be able to make given this estimated demand. In some cases the problems may be the same, but they don't have to be the same in general.
    
    The general challenge of forecasting is typically formulated using an auto-regression modeling framework. The goal is to use current or historical demand or sales information to predict future sales or demand. Thus, most forecasting problems are supervised prediction problems. The recent survey by Cadavid, Lamouri, and Grabot outlines some of the trends and advances in solving these business problems using prediction and pattern mining algorithms \cite{sdforecasting_usugacadavid:hal-01881362}. Their paper discusses predicting future sales and demand values using a combination of past demand or sales values, endogenous variables like price, and exogenous variables like day of week. They go on to outline the standard supervised modeling methods for performing regression and classification including neural networks, random forests, support vector machines, and many other kinds of models.
    
    It should be noted that in section 2.3 of the survey on sales and demand forecasting describing 'prescriptive analytics' methods for prescribed improvements is statistically incorrect. Using feature importance measures to craft action intervention suggestions to change a predicted value is statistically invalid (The book 'All of Statistics' \cite{Wasserman:2010:SCC:1965575} has a good example of counterfactual analysis that demonstrates this statistical fallacy). It is a classic case of miss-applying statistical associations to answer causal questions (The old 'correlation is not causation' trap). A main point of our paper is to identify the correct methods for answering causal questions and for prescribing intelligent actions in statistically valid ways.

\subsubsection{Stock Market Analysis}

    Investing in companies in the stock market is all about predictions. Humans have been making predictions about the stock market since it's inception. The goal is for the investor to trade money to a company in exchange for stocks in a company, in the hopes that the company will use that money to grow the business and increase the overall value of that company's stock. Humans make many predictions about stocks. Humans evaluate if the current price of a stock is an accurate representation of it's value, and make predictions about whether or not a stock's value will increase or decrease. Prediction and pattern mining methods are applied to answer these questions as well. This paper \cite{Shah_2019} by Shah, Isah, and Zulkernine creates a taxonomy of prediction and pattern mining methods. They also identify macroeconomic analysis, industry analysis, and company analysis as three major areas of stock market analysis where these methods are applied. In another survey \cite{smtrading_ml_2016}, Li, Wu, and Bu discuss machine learning methods for quantitative trading specifically. They survey many machine learning methods for developing trading strategies, predicting metrics like mean reversion of stock price, and understanding characteristics of stock price behaviors. Other groups have gone so far as to incorporate social media posts into the stock value prediction problem \cite{twitter_stock_market_ml_2016}.

\section{Causality}

    In this section, we delve into the underutilized area of causal analysis (otherwise referred to as causal inference). Causal analysis is a relative newcomer in terms of having a formal mathematical language for performing analysis. While the ideas around causality can be traced back to discussions the Greeks had on the topic, naming it 'Etiology', much of the formal mathematical work in causal inference was developed by research performed by Ronald Fisher, Jerzy Neyman, Judea Pearl, Donald B. Rubin, Paul Holland, and many others. A good summary of the history of causal inference and some of its more modern foundations can be found here \cite{FOUNDATIONS_OF_CAUSAL_INFERENCE_JP_2010} \cite{Pearl:2018:BWN:3238230}. Other good references on the topic of causality include \cite{pearl2016causal_primer} \cite{Pearl:2009:CMR:1642718} \cite{Wasserman:2010:SCC:1965575} \cite{morgan_winship_2014} \cite{Peters:2017:ECI:3202377}

    We start the section by discussing causal analysis in the area of controlled experimentation. Namely, we discuss basic concepts from the field of experimental design and how experiments are typically constructed to estimate the possibility of cause and effect. The method in business called A/B testing (the applied business nomenclature of general hypothesis testing) as a means to understanding the cause and effect relationships of making a particular decision. We also introduce some mathematical notation of experiments from \cite{Holland:10.2307/2289064}, and specify some of the assumptions inherent to most scientific methods of experimentation. 
    
    Then, we discuss less than ideal experimental conditions with the topic of Quasi-experiments. Quasi-experiments are experiments in which a researcher may have the ability to assign treatment or control, but cannot do so completely randomly. It deals with issues of what is known as internal validity. Internal validity is a measure of how well a study can rule out alternative explanations for findings. Examples of a violation of internal validity include confounding and selection bias. Even random assignment may not guarantee that both groups will be constructed in a way that their effects of treatment are from comparable baselines. We discuss the basic setup of Quasi-experiments, and methods for estimating cause and effect including difference in differences, regression discontinuity design, and uplift modeling.
    
    Finally, the topic of causal analysis in observational studies is presented. The main methods described here include the potential outcomes framework, as well as causal graphical models. We briefly touch on how these methods can be used to perform the tasks of causal discovery, causal identification, and causal estimation.
    
    In the applications section, we discuss the main fields where causal analysis is being applied, including the social sciences, medicine, and much more recently, the area of customer experience. While the social sciences and medicine have enjoyed many years of applying and advancing methods of causal analysis, businesses in other areas of industry have been lagging behind or neglectful of the area of causal analysis. 

\subsection{Controlled Experiments}

\subsubsection{Origins of Causal Philosophy}

    Identifying Cause and Effect is at the heart of nearly all scientific endeavors. Most scientific inquiry is focused on why things act the way they do, and how they came about. A causal understanding of environments and actions allows us to explain the world, build complex systems, and construct strategies for bringing about desired change. Some of the earliest recorded materials on cause and effect estimation come from the ancient Greek philosophers, with Aristotle outlining the material, formal, efficient, and final causes as the 4 fundamental causes for a thing. Some other philosophers with important contributions to the idea of causality include David Hume, John Stuart Mill, and Patrick Suppes \cite{Holland:10.2307/2289064}. Most of the determination of cause and effect in the last few thousand years has been done through experiments. Experiments are repeatable procedures constructed to accept or refute a hypothesis. In our case, we are referring to experiments designed to support or refute a causal hypothesis. As we'll discuss, experiments involving causal conclusions must include some assumptions about the experiment, or nature of a cause and effect relationship. Many times in classic experiments these assumptions are implicitly defined and accepted in the scientific community for which experiments are conducted. For our purposes, we will present notation introduced by Paul Holland \cite{Holland:10.2307/2289064} that will allow us to formalize causal estimation mathematically, and to clearly identify the assumptions of causal effect estimation methods.
    
\subsubsection{Notation for Causal Effect Estimation}

    We directly use Paul Holland's discussion of causal analysis under the Neyman-Rubin (Potential Outcomes) causal model as a baseline for modeling causality throughout this section \cite{Holland:10.2307/2289064}. The potential outcomes model has been found to be a useful tool for formalizing and describing causal assumptions in traditional experimental, as well as observational settings. We start by defining the idea of a cause or treatment. A treatment is a cause, or stimulus that is applied to some unit in an environment. A treatment or cause is defined in relation to another cause. In experimental setups, a treatment is defined in relation to a control group. Thus, there are two causes that are defined (in this case, the causes are treatment applied and treatment not applied or control). Next, we define a population of units, \(U\). Units are the entities in an environment upon which a treatment acts. A treatment is either applied to a unit, or not applied. More specifically, let's assume there are two levels of treatment, \(t\) (treatment) and \(c\) (control). Then either \(t\) is applied to unit \(u\), or \(c\) is applied to unit \(u\). Let's further define a variable \(S\) to be a variable representing the cause to which each unit in \(U\) is exposed. Then, \(S = t\) means the unit is exposed to \(t\) and \(S = c\) means the unit is exposed to \(c\). With these definitions, we've defined not only the idea that a unit is exposed to a treatment, but also the idea that there could have been an alternative exposure. In other words, any unit is potentially exposable to a treatment, and given a set treatment, there is an alternative treatment that could have been made. Finally, we define \(Y\) as the response or outcome variable. The role of a response variable \(Y\) in association analysis is to represent the observed value of a response. However, in causal analysis (under the potential outcomes model) two response variables are needed, \(Y_t(u)\) and \(Y_c(u)\). \(Y_t(u)\) represents the value of the response that would be observed if unit \(u\) was exposed to treatment \(t\) and \(Y_c(u)\) represents the value observed on the same unit if that unit was to be exposed to \(c\). Given this definition, the unit causal effect is defined as:
    
    \begin{center}
        t causes the effect
        \begin{equation} \label{eq2}
            Y_t(u) - Y_c(u)
        \end{equation}
    \end{center}
    
    This gives rise to what Holland describes as the Fundamental Problem of Causal Inference. He states that it is impossible to {\it observe} the value of \(Y_t(u)\) and \(Y_c(u)\) on the same unit, and therefore, it is impossible to {\it observe} the effect of \(t\) on \(u\). Either a treatment is applied or it is not. Therefore, only one of the effects can actually be observed (unless some assumptions are made, which are outlined in the next sub section).
    
    If homogeneity or invariance assumptions are not made on the relationship between \(Y_t(u)\) and \(Y_c(u)\), then a more statistical approach to measuring cause and effect is used. In this approach, rather than attempting to measure the specific cause and effect of a unit, the average causal effect \(T\) is calculated as:
    
    \begin{center}
        \begin{equation} \label{eq2}
            E[Y_t - Y_c] = E[Y_t] - E[Y_c] = T
        \end{equation}
    \end{center}
    
    This equation shows that information on different units that can be observed can be used to estimate information about the average causal effect \(T\). As Holland points out, 'The important point is that the statistical solution replaces the impossible-to-observe causal effect of \(t\) on a specific unit with the possible-to-estimate average causal effect of \(t\) over a population of units'. Given these definitions of causal effect and average causal effect, we can now shed light on some causal assumptions from classic experimentation setups.

\subsubsection{Traditional Causal Assumptions}

    \begin{itemize}
        \item {\bf Temporal Stability and Causal Transience}:  If we assume that (a) the value of \(Y_c(u)\) doesn't depend on when the sequence of treatment exposure and measurement occurs and (b) the value of \(Y_t(u)\) is not affected by the prior exposure of \(u\) to the treatment from (a), then \(Y_c(u)\) and \(Y_t(u)\) can simply be measured by successively applying treatments and measuring the causes at differing points in time. This is analogous to turning on and off a light switch to determine the effect of the switch. We assume that the flip of the switch now has no affect on the outcome of the later flip of the switch, and that everything else remains the same. This is a commonly applied assumption in both science and every day life.
        \item {\bf Unit Homogeneity}: If we assume that \(Y_t(u_1) = Y_t(u_2)\) and \(Y_c(u_1) = Y_c(u_2)\) for two comparable units \(u1\) and \(u2\), then we can take the causal effect of \(t\) as \(Y_t(u_1) - Y_c(u_2)\). In science, comparable units are typically selected by the scientist for comparison. This too is a commonly applied assumption in science and every day life.
        \item {\bf Independence}: Most classic experiments rely upon randomization of applying a treatment to units to create independence. This is important because the only information over the observed data set that can be calculated is \(E[Y_s | S = t] = E[Y_t | S = t)\) and \(E[Y_s | S = c] = E[Y_c | S = c)\). This is clearly the case because we can only observe the value of the outcome variable corresponding to the application of a particular treatment. We can't physically observe the hypothetical alternative outcome given a treatment. However, if we can apply randomized application of treatments, then by the laws of probability and independence \(E[Y_s | S = t] = E[Y_t])\) and \(E[Y_s | S = c] = E[Y_c])\). This is in general not the case, and only holds when the outcome is independent of the treatment assignment. Given this, it can be derived that the causal effect is \( T = E[Y_t] - E[Y_c] = E[Y_s | S = t] - E[Y_s | S = c] = E[Y_t | S = t] - E[Y_c | S = c]\). This is called the {\it prima facie causal effect} of \(t\) relative to \(c\) and can be denoted \(T_{PF}\)
        \item {\bf Constant Effect}: The final typical assumption is one of constant effect. This assumption is that the effect of \(t\) on every unit is the same. Given this assumption, the causal effect can be stated as \(T = Y_t(u) - Y_c(u) \forall u \in U \). Stated another way, the treatment \(t\) adds a constant amount \(T\) to the control response for each unit. This assumption in general does not hold for practical experiments, especially in the cases where the environment and treatments have complex causal interactions requiring complex models.
        \item {\bf Stable Unit Treatment Value Assumption (SUTVA)}: This is the assumption that the potential outcome value of one unit is not affected by the assignment of treatments to other units. This is a sort of independence assumption between the experimental units. An open research area is to understand how to quantify causal effect when the SUTVA assumption doesn't hold.
    \end{itemize}

\subsubsection{Design of Experiments}

    Given our previous definitions of causality and the traditional assumptions made, we now briefly touch on experiments and experimental design. An experiment is an empirical method for providing support for a pre-specified hypothesis or for rejecting the hypothesis by counterexample. Experiments many times have an independent variable that can be manipulated or affected by the experimenter, and a dependent variable that is measured. Any experiment must be constructed so that it is repeatable, with clearly interpretable results that can be trusted. To ensure that the experiment can be trusted, an experimenter must develop their experiment to control for sources of bias or spurious association (also called confounders). Classic controlled experiments attempt to control for confounding through both randomization and control of the experimental units and their environment. An experimental designer must also concern themselves with internal and external validity of the experiment. Internal validity is a measure of how well an experiment can rule out alternative explanations for its conclusions. Issues of internal validity arise due to things like confounding, selection bias. External validity is a measure of how much the conclusions of an experiment can be applied outside of that experiment. Stated another way, it's how valid the conclusions are to contexts outside (external) to the experiment. The goal of experimental design is to construct an experiment that has high internal and external validity.
    
    There are a few types of experiments worth noting. The first is called a controlled experiment. A controlled experiment is an experiment where experimental sample units are compared against a control set of sample units. Experimental sample units are units for which some treatment or manipulation of an independent variable is applied, and control sample units are units for which some treatment or manipulation is withheld. The basic assumptions of this experiment type is that treatments are randomly assigned (resulting in the independence assumption) and that the control is such that the units in the treatment and control groups are comparable on average and without bias (an assumption implicit in the construction of \(Y_t(u)\) and \(Y_c(u)\) for potential outcomes and similar to the unit homogeneity assumption). 
    
    The second major kind of experiment is called a natural experiment or quasi-experiment. These experiments are much more practical in complex and uncontrolled environments because the assumptions maintained for controlled experiments are almost never realistic in environments other than a laboratory. Quasi-experiments focus solely on observations of the variables of a system under study. They differ from controlled experiments because they focus on the collection and analysis of data in a way that allows for the effects of variables to be determined. Quasi-experiments mainly focus on attempting to remove spurious associations and sources of bias from the data collected. The challenge here is that unobserved variables that aren't held constant can create uncontrollable associations, leading to inaccurate conclusions. 
    
    The final kind of experiment we introduce is called an observational study. In fact, an observational study is not truly an experiment as the data collection process does not follow from the construction of an experiment. Instead, observational studies are analyses over data collected in an observed system in which the independent variable isn't under the control of the experimenter. In a quasi-experiment, an experimenter may still be able to control the assignment of treatment to patients, albeit in a less than random way. In observational studies however, there is no real control of assignment of treatment, and no notion of control over the process in which the data were collected. The distinction between quasi-experiment and observational study is somewhat blurred in their definitions. We make the distinction that while quasi-experiments still have a level of control as to assignment of units to a treatment or control group, the application of treatment within these groups is not totally random in the sense of the universe of possible experimental units. Thus, quasi-experiment must control for this systematic bias still. Observational studies not only have non-random assignment over the treatment, but also no control over the assignment of subjects into a treated group versus a control group. Another way to think of this is in terms of pre/post treatment. In observational studies, units are assigned to treatment and control groups post-treatment. In quasi-experiments, units are assigned to treatment and control groups pre-treatment. In the practical setting of causal analysis, most of the data available is observational. After causal analysis has been done in the observational setting, actions can be applied in a quasi-experimental fashion to systematically estimate their effect.

    Now that we've identified the 3 main kinds of experiments, we consider a classic example of an experimental setup used often in industry to test a hypothesis. This setup is called A/B testing. AB testing is widely considered go-to methodology for testing the impact of a change in a software product. Users are usually split into a control group where users interact with software containing the original feature, and and experimental group with users who interact with software containing the new or updated feature. Through a randomized experiment (in reality, it should be treated more as quasi-experimental), A/B testing is used to estimate the causal impact of the treatment. A/B testing consists of the following steps:
    
    \begin{enumerate}
        \item {\bf Decide the Metrics}: Choose a high-level metric that we want to focus on (active users, probabilities and rates, ratios).
        \item {\bf Determine a Hypothesis}: Choose a hypothesis that relates an an independent variable (a variable which is manipulated by the experimenter) and the dependant variable.
        \item {\bf Sample}: Determine how to choose samples from the population. Sampling is usually done uniformly random, but there are often confounders that need to be controlled for in an experimental manner.
        \item {\bf Choose Significance Level}: If the hypothesis is one about a statistical association measure, then a choice of level (alpha) and statistical power (1-Beta) are chosen. These measures are used to make the final decision of whether to reject the hypothesis or not.
        \item {\bf Run Test}: Given the setup, the A/B test is then run over a time period where the treatment and control groups are simultaneously treated (or not treated) and the important metrics are collected.
        \item {\bf Analyze and Conclude}: Given the final metrics, use a statistical test to determine whether the hypothesis can be rejected or not. In the following sections, we'll show how association based measures can be transformed into causal conclusions. We'll note here however that data and statistics alone isn't sufficient to make causal conclusions, and there is always a certain set of causal assumptions that must be made in order to derive causal conclusions. Causal analysis concerns itself with what these assumptions are, how to validate them, and how to use the result of causal assumptions to estimate causal effects given a data set.
    \end{enumerate}

\subsection{Quasi-Experiments}

    In the previous 'Design of Experiments' subsection, we outlined three major environments in which we want to perform causal analysis. In the previous section, we outlined controlled experiments as a classic method for estimating cause and effect. However, industry applications of causal analysis are done in either a quasi-experimental or an observational manner. In this sub section, we outline the setup of a quasi-experiment and detail the methods for estimating causal effect. We briefly describe Difference in Differences, Regression Discontinuity Design, Propensity Scoring, and Uplift Modeling as methods for estimating causal effect in quasi-experimental setups.

\subsubsection{Difference in Differences}

    The Difference in Differences (DID) method attempts to estimate the causal effect of a treatment through the differential comparison of effects in a treatment and control group in an uncontrolled experiment that has potential confounding and bias issues. It estimates causal effect by comparing the average change over time in an outcome variable between the treatment and control group. A DID experiment requires data measured from a treatment group and a control group at two times. Variables in the control and treatment must be measured at one time before and one time after treatment and non-treatment. The reason for a measurement before and after treatment is to estimate the causal effect due only to the treatment. This is because the control and treatment groups may not have started at the same initial conditions in the experiment. By taking the measurements for both groups over time, the DID method can calculate change not due to the treatment (by total change in the control group) along with change due to the treatment (change in treatment group, minus total comparable baseline change from the control group). 
    
    The method starts by breaking down the experimental units into two groups. The groups are the treatment group with \(S=t\) and the control group with \(S=c\). DID experiments also define two time periods, defined by the variable \(P\). DID defines \(P=0\) as the pre-treatment period and \(P=1\) as the post-treatment period. Given this, we can define the outcome variables in the potential outcomes framework notation as \(Y_{S}(u, p)\) where \(Y_{t}(u, p)\) is the potential outcome unit \(u\) attains in period between \(P=0\) and \(P=1\) with treatment \(S=t\). We can also define \(Y_{c}(u, p)\) is the potential outcome unit \(u\) attains in period between \(P=0\) and \(P=1\) with treatment \(S=c\). In this case, we can define the following:
    
    \begin{itemize}
        \item The Causal Effect for Unit \(u\) at Time \(p\) is:  
            \begin{center}
                \begin{equation} \label{eq2}
                    \tau(u, p) = Y_{t}(u, p) - Y_{c}(u, p)
                \end{equation}
            \end{center}
        \item The Observed Outcomes Are: 
            \begin{center}
                \begin{equation} \label{eq2}
                    Y(u, p) = Y_t(u, p) - Y_c(u, p)
                \end{equation}
            \end{center}
        \item The Fundamental Problem of Causality: If we are calculating treatment effect at time \(p=1\), then we want to estimate \(Y(u, p=1) = Y_t(u, p=1) - Y_c(u, p=1)\), but only one of the potential outcomes can actually be observed at that time.
        \item Average Effect of the Treatment on the Treated (ATT): 
            \begin{center}
                \begin{equation} \label{eq2}
                    \tau_{ATT} = E[Y_t(u, p=1) - Y_c(u, p=1) | S = t]
                \end{equation}
            \end{center}
        
        \item Identification Assumption (parallel trends assumption):
            \begin{center}
                \begin{equation} \label{eq2}
                    E[Y_c(u, p=1) - Y_c(u, p=0) | S = t] = E[Y_c(u, p=1) - Y_c(u, p=0) | S = c)
                \end{equation}
            \end{center}
        \item Full ATT Given the Identification Assumption is: 
            \begin{center}
                \begin{equation} \label{eq2}
                    \begin{split}
                        E[Y_t(u, p=1) - Y_c(u, p=1) | S = t] = \\
                        (E[Y_t(p=1) | S=t] - E[Y_c(p=1) | S=c]) \\
                        - (E[Y_t(p=0) | S=t] - E[Y_c(p=0) | S=c])
                    \end{split}
                \end{equation}
            \end{center}
    \end{itemize}
    
    These definitions, assumptions, and derivations characterize the difference in differences method for determining the causal effect of a treatment. However, a visual representation can shed more light on the setup and the components of the ATT equation above. The figure below characterizes the main components of the ATT equation calculation. The diagram below maps to the ATT equation in the following way:
    
    \begin{center}
        \begin{equation} \label{eq2}
            \begin{split}
                P_1 &= E[Y_t(p=0) | S=t] \\
                S_1 &= E[Y_c(p=0) | S=c] \\
                P_2 &= E[Y_t(p=1) | S=t] \\
                S_2 &= E[Y_c(p=1) | S=c] \\
                Q &= P_1 - S_1           \\
                ATT &= (P_2 - S_2) - Q   \\
            \end{split}
        \end{equation}
    \end{center}
    
    \begin{figure}[h]
      \centering
        \includegraphics[width=8cm, height=8cm]{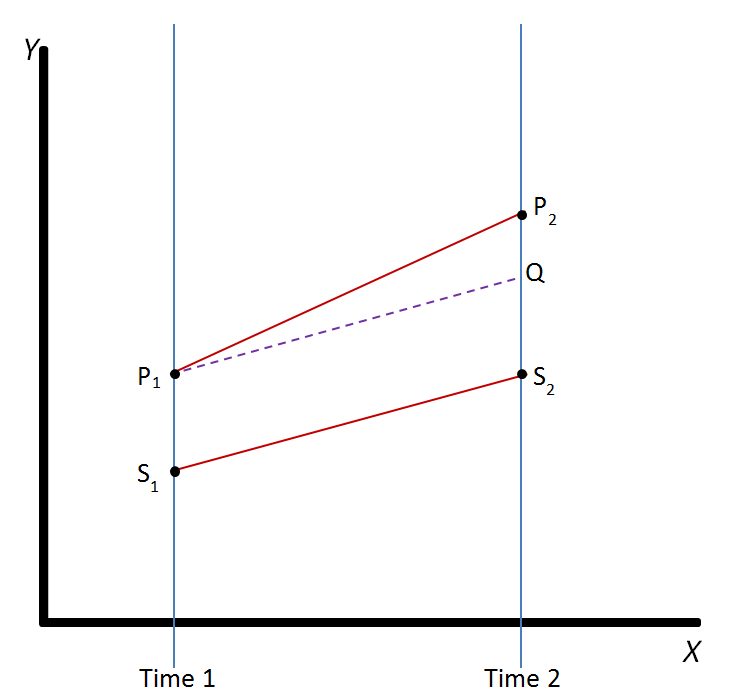}
      \caption{Illustration of the Difference in Differences Method \cite{ruthvan_2014}}
    \end{figure}
    
    It's clear from the figure that the calculation of causal effect of the treatment is simply a method to remove the baseline bias between the control and treatment groups. It's also clear how the parallel trends assumption factors into the estimation, which is illustrated by the dashed line to \(Q\). Given our definitions and assumptions, we can calculate the ATT from a data set by using the sample means as estimators of the expectations for the observed data in the definition of ATT given the Identification Assumption. An estimator for pooled samples can also be calculated through the use of a linear regression model. While the DID method is highly aligned with classic experimental design, it's applicability to practical problems is limited. Michael Lechner provides an in depth treatment of the DID method here \cite{DID_Depth:usg:dp2010:2010-28}. More information on the practical use of the DID method can be found here \cite{DID_Trust:10.1162/003355304772839588}.

\subsubsection{Regression Discontinuity Design}

    Regression Discontinuity Design (RDD) is a quasi-experimental setup that is used when a threshold is used to determine the assignment of a treatment. Experimental units are assigned to a treatment group if they lie above a threshold, and are assigned to a control group if they lie below a threshold. The causal effect of a treatment is then measured as the magnitude of the average discontinuity above and below this threshold for the treatment group versus the control group. The basic idea is to compare local treatment effects around the threshold used for assigning treatment. The causal assumption here is that units near to the threshold are sufficiently similar as to be comparable for estimating the average causal effect of the treatment.
    
    An example of this is one of the original applications of RDD to evaluating the causal effect of merit based scholarships \cite{Thistlethwaite1960RegressiondiscontinuityAA}. In this context, the goal is to understand the effect of providing merit-based scholarships on academic performance. The issue is that students that are already high performing are likely to continue to be high performing, even after being assigned a scholarship. There is an analogous case for lower performing students. Thus, there is an inherent bias in the assignment of treatment that does not follow a classic random assignment methodology.  
    
    The most common approach to RDD estimation is a non-parametric method using local linear regression on either side of the threshold for computing and average treatment effect measure. The major components of the analysis are:
    
    \begin{itemize}
        \item Non-Parametric Regression: 
            \begin{center}
                \begin{equation} \label{eq2}
                    Y_S = \alpha + \tau*D + \beta_1(X_S-l) + \beta_2*D(X_S-l) + \epsilon
                \end{equation}
            \end{center}
            where \(D=1\) if \(S=t\) and \(D=0\) if \(S=c\), \(l\) is the threshold or level cutoff for the treatment assignment.
        \item Regression Bandwidth: \(h\) is the bandwidth of the kernel used to fit the regression such that
            \begin{center}
                \begin{equation} \label{eq2}
                    c - h <= X <= c + h
                \end{equation}
            \end{center}
        \item Average Treatment Effect Calculation (ATE):
            \begin{center}
                \begin{equation} \label{eq2}
                    \begin{split}
                        ATE &= E[Y_t | S=t ] - E[Y_c | S=c] \\
                            &= E[Y_t | S=t, u=threshold ] - E[Y_c | S=c, u = threshold] \\
                            &= Y_t(u=threshold) - Y_c(u=threshold)
                    \end{split}
                \end{equation}
            \end{center}
            where \(Y_S\) in this case is the value of the outcome variable defined by the non-parametric regression. 
    \end{itemize}
    
    The expected value regression at u=threshold uses a 'local unit homogeneity' causal assumption to calculate the average treatment effect. It is assumed that at this local point, the difference between the expected outcome for treatment and control is comparable and representative of effect estimation. A more in-depth treatment of the RDD method can be found here \cite{Thistlethwaite1960RegressiondiscontinuityAA}.

\subsubsection{Uplift Modeling}

    Uplift modeling is the last method we mention in the area of quasi-experimental design. Uplift modeling uses complex statistical and machine learning based methods along with quasi-experimental design to directly predict the effect of an intervention on a particular experimental unit, given the experimental units are not all comparable, and the assumption of constant treatment effect does not hold (aka, treatment has a different effect given different attributes of an experimental unit). Uplift modeling relies on quasi-experimental design to predict the incremental response in an outcome variable due to a single action.
    
    The basic setup of the problem of uplift modeling aligns with the notion of potential outcomes and the use of average treatment effect. The conceptual goal is to train a predictive model for the treatment group and control group in turn, and to model the predicted incremental effect of the action as the difference in these predictions. Specifically, we define the following:
    
    \begin{center}
        \begin{equation} \label{eq2}
            \begin{split}
                f_1 = E[Y_i | X_i, S=t] = f(X_i, D_i, X_i, X_i*D_i) \\
                f_2 = E[Y_i | X_i, S=c] = f(X_i, D_i, X_i, X_i*D_i) \\
                T_i = E[Y_i | X_i, S=t] - E[Y_i | X_i, S=c] = f_1 - f_2 \\
            \end{split}
        \end{equation}
    \end{center}
    
    where \(D_i=1\) if \(S=t\) and \(D_i=0\) if \(S=c\), \(T_i\) is the individual treatment estimate for experimental unit i, and \(f_1\) and \(f_2\) are the statistical or machine learning models trained on the subset of data treated versus not treated. The main approaches to uplift modeling are typically tree based methods that directly incorporate information from treatment and control groups into the tree splitting criteria during training, and regression models that split out regression coefficients to coefficients for prediction across treatments, and coefficients within treatments (which is very similar to the idea of parameter based transfer learning, where the shared regression coefficients are treatment/domain agnostic). More specific information on uplift modeling can be found in the following references \cite{True_Lift_Modeling} \cite{Radcliffe2007UsingCG}.
    
\subsection{Observational Studies}

    Observational studies are where the use of causal analysis methods begin to shine in industry environments. It is more typical in the data science realm to receive observational data that was not collected under an experimental design. The two main methods for performing causal analysis in the observational domain are the potential outcomes framework, and the causal graphical model framework. While there are other possible models of causality (such as Grainger causality), these two methods are used much more in practice, and allow for more reasonable and realistic causal conclusions. 

\subsubsection{Potential Outcomes}

    Throughout this section, we have described experimental setups using the notation mainly of potential outcomes. This notation is defined in the 'Notation for Causal Effect Estimation' subsection. We refer the reader back to this section for a review on the notion that defines the potential outcome framework. We re-iterate here three critical components of the potential outcome model. The first component is the fundamental problem of causal inference. This problem is that we cannot observe the value of both \(Y_t(u)\) and \(Y_c(u)\) simultaneously. Only the value of the variable corresponding to the treatment of the unit can be observed. The second component is the idea of 'no causation without manipulation'. Paul Holland and others have proposed the idea that the notion of causal effect estimation only makes sense in the context of manipulation. Thus, under their original proposition, we can estimate the effect of only things we believe we could manipulate. As an example, it does not make sense to attempt to quantify the effect of genes on the occurrence of a disease because genes are not something that can be manipulated by a researcher. However quantifying the effect of a drug on the occurrence of a disease is estimable because we can manipulate the application of a drug. Thus, attributes of an experimental unit that are not manipulable cannot have their causal effect estimated (Because the analysis results in an association statement mathematically, not a causal statement). This leads to some confusing conclusions (such as genes don't cause diseases) which have been debated amongst causal inference researchers. The next section on causal graphical models defines causality in such a way that the causal effect of attributes can be estimated. The third component of potential outcomes is the SUTVA assumption. This isolation assumption allows for the estimate of a potential outcome independent of the treatments and potential outcomes of other units.

    While the notation of potential outcomes specifies the mathematics of causal reasoning, by itself it doesn't suggest a method for performing estimation. We showed before that under specific assumptions we can perform a causal estimation, but we would prefer a more principled method of estimating causal effect. Propensity score matching (PSM) is a very general method of handling observational study setups when the assumption of random assignment to treatment our control group is not random or under the control of an experimenter. It's a method that is often used under the framework of potential outcomes for performing general effect estimation. It quantifies causal effects by attempting to account for the covariates in the data set that predict being assigned treatment. Stated another way, the method attempts to control for potential sources of confounding by identifying the source variables and adjusting for them. In randomized experiments, the idea is that the randomization will cause the covariates to be balanced across treatment and control groups on average. The goal of PSM is to take the sample of units that received treatment, and to construct a sample of control units that match the treatment sample on all observed covariates. By doing this, PSM creates the sample comparisons as effectively randomly sampled. This method is particularly useful for not only quasi-experiments, but also for observational studies. PSM is typically used when treatment and control groups are not comparable, and it's challenging to select a subset of comparison units because the data is high dimensional and it's not clear how to create or identify comparable units across the groups. PSM works by predicting the probability of a unit belonging to either the treatment group or the control group.

    Before describing the matching procedure used for causal estimation, we describe some of the major components and assumptions of the model. These include:
    \begin{itemize}
        \item {\bf Strong Ignorability Assumption}: This assumption is key to being able to rely on the causal estimates of matching. This assumption states that assignment to the treatment or control group is strongly ignorable if the potential outcomes are independent of the treatment, conditional on the background variables. That is, \(Y_t, Y_c \perp S | X \).
        \item {\bf Balancing Score}: This score, \(b(x)\), is used to estimate the quality (or weighting) of the matching of control to treatment units such that \(S \perp X | b(x)\). The score measures the ability to distinguish or predict which group (control or treatment) a unit belongs to given the covariates. If this is not possible (because the treatment is independent of the covariates given the score/weighting), then the matching is good and the bias between the groups has been removed.
        \item {\bf Propensity Score}: The propensity score is the direct, baseline measure of the ability to predict or assign an experimental unit to a treatment group. Specifically, \(e(X) \coloneqq Pr(S = t | X)\). Notice that this is an estimate of the probability of a unit belonging to the treatment group. If the unit is from the control group, then the propensity score on that unit estimates the similarity of the control unit to the sample of treatment units. Thus, propensity is effectively a measure of transfer that calculates the similarity of the control unit to units in the treatment group, even though it doesn't belong to the treatment group.
    \end{itemize}

    Given these definitions, we can summarize the main theorems of the method as follows \cite{PSM_Orig:10.2307/2335942}:
    \begin{itemize}
        \item The propensity score \(e(X)\) is a balancing score \(b(x)\).
        \item Any score 'finer' than the propensity score is a balancing score, or \(e(x) = f(b(X))\).
        \item If \(Y_t, Y_c \perp S | X\) then the following are true:
            \begin{itemize}
                \item \(Y_t, Y_c \perp S | e(X)\)
                \item For any value of balancing score, the difference in sample treatment averages,\(\bar{Y_t} - \bar{Y_c}\) over units with the same balancing score, is an unbiased estimate of average treatment effect \(ATE = E[Y_t] - E[Y_c]\).
            \end{itemize}
        \item Estimates of the balancing score of sample units can produce a sample balance on \(X\).
    \end{itemize}

    The basic procedure for PSM is as follows:
    \begin{enumerate}
        \item {\bf Build a Predictive Model of Propensity Score}: Build a predictive model \(Pr(S_i|X_i)\) (for example a logistic regression or other machine learning model) that predicts the value of \(S_i = t or c\) (propensity score) for unit \(i\).
        \item {\bf Check Propensity Score Balance}: Compare the distributions of propensity score in the treatment group and the control group. If the distributions look significantly different (meaning there isn't much overlap in covariate values) in shape within strata, then the final matching and estimate may be biased. 
        \item {\bf Match}: Match each unit in the treatment group with one unit from the control group based off of propensity score. There are many methods for performing this match including nearest neighbor matching, caliper matching, mahalanobis distance, stratification matching, difference in differences matching, exact matching, and genetic matching (which generalizes propensity matching and works well in practice).
        \item {\bf Verify Covariate Balance}: Examine the covariate distributions across the treatment and control groups to ensure covariate balance. Jasjeet Sekhon discusses some of the challenges in using propensity score matching and how sometimes in practice, more basic matching methods can actually result in further imbalance in covariates than the original data set.
        \item {\bf Perform Causal Analysis}: Calculate the sample average of the difference in sample treatment averages,\(\bar{Y_t} - \bar{Y_c}\) over the paired units, and use this as an unbiased estimate of average treatment effect \(ATE = E[\bar{Y_t} - \bar{Y_c}\).
    \end{enumerate}
    
    The potential outcomes framework is very popular for reasoning about causal estimation, and has a wide variety of methods that allow for causal estimation in observational settings. More information on the potential outcomes framework and its connection to other causal inference methods can be found in the following references \cite{PSM_Orig:10.2307/2335942} \cite{Wasserman:2010:SCC:1965575} \cite{morgan_winship_2014}.

\subsubsection{Causal Graphical Models}

    Causal graphical models are a relative newcomer to the field of causal analysis. Causal graphical models (CGM) draw their inspiration from bayesian network models, as well as from structural equation models (SEM). This is no surprise as the development of both bayes nets and causal graphical models are largely due to the efforts of Judea Pearl. We delve into the specifics of bayes nets later in this article in section 7.2.2. For a more in-depth treatment of the topic, we refer the reader to \cite{Russell:2009:AIM:1671238}. In this section, we pull largely from the in depth works by Judea Pearl \cite{pearl2016causal_primer} \cite{Pearl:2009:CMR:1642718} \cite{Pearl:2018:BWN:3238230}.
    
    The basic idea of causal graphical models is that cause and effect relationships can be estimated using a causal graphical model, a set of statistical adjustments, and a set of data. The graphical model consists of variables, and directed arrows between variables which represent a cause and effect relationship. The graphical model is a directed acyclic graph (DAG), where causality is defined as when one outcome variable \(Y\) assigns it's value based on the value of a cause variable \(X\) (pearl describes this as 'Y listens X when determining its value') \cite{pearl2016causal_primer}. This is explicitly represented in the relationship between structural equation models and causal graphical models, where each variable in a graph is assigned an equation that depends on it's parents. For example, given the causal graphical model below, we can identify \(X\) and \(Y\) as exogenous variables, \(Z\) as an endogenous variable, and a set of structural equations \(F={f_Z}\). Then, we can define a causal structural equation model that aligns with this causal graphical model as \(f_Z : Z = \alpha X + \beta Y\) meaning that the value of \(Z\) is set in reaction to (note this meaning beyond 'equality') the combination of \(\alpha X + \beta Y\). 
    
    \begin{figure}[h]
        \begin{center}
            \includegraphics[width=3cm, height=3cm]{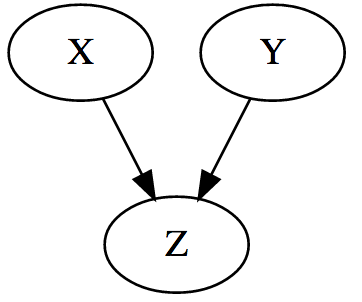}
            \caption{}
        \end{center}
    \end{figure}
    
    The major steps for performing causal analysis using CGMs are:
    \begin{enumerate}
        \item {\bf Causal Discovery}: Here, the goal is to determine the structure of the graph (potential edges), as well as edge directions. While some edges and directions can be inferred by data, data alone can only identify up to the level of equivalence classes of causal graphs on data. There also exists a fundamental 'causal assumption' which is to assume that the directed arrows represent cause-effect relationships as defined for this type of analysis. 
        \item {\bf Causal Identification}: Here, the goal is to use the assumed causal graph to determine whether or not a cause and effect relationship can be determined between two variables in the graph (potentially in the presence of unobserved confounders). 
        \item {\bf Causal Estimation}: If a cause and effect relationship can be estimated, then we need to generate a set of adjustment variables that will allow us to use classic statistical estimation methods to estimate the cause and effect relationship. Two major criteria for doing this in CGMs are the back-door and front-door criteria. Given a final adjustment formula for estimating cause and effect, classical statistical methods are used to make the final computation.
    \end{enumerate}
    
    The fundamental structures of a graph are chains, forks, and colliders. The figure below provides these fundamental graphical structures. We do not give a full treatment of these structures here, and instead refer the reader to the figure below and the following resources \cite{pearl2016causal_primer} \cite{Pearl:2009:CMR:1642718}. We will however point out some important characteristics of chains and colliders. First, if we have the chain structure in the figure below, then \(Z\) is independent of \(X\) given \(Y\), \(Z \perp X | Y \). Second, if we have the fork structure in the figure below, then \(Z\) is independent of \(X\) given \(Y\), \(Z \perp X | Y \). Finally, if we have the collider  structure in the figure below, \(X\) is dependant on \(Y\) given \(Z\) or any children of \(Z\), \(X \not\!\perp\ Y | Z \).

    \begin{figure}[htp]
        \centering
        \label{figur}\caption{Fundamental causal graph structures}
        \subfloat[Chain]{\label{figur:1}\includegraphics[width=1.5cm, height=5cm]{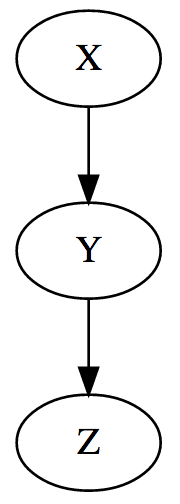}}
        \subfloat[Fork]{\label{figur:2}\includegraphics[width=3cm, height=3cm]{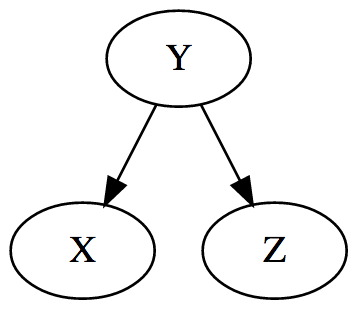}}
        \subfloat[Collider]{\label{figur:3}\includegraphics[width=3cm, height=5cm]{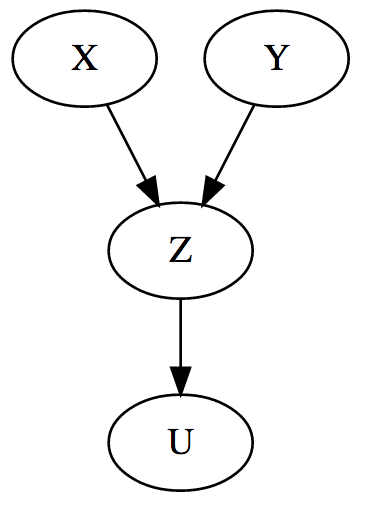}}
    \end{figure}
    
    These structures in the graph are important because they are used to define an important criteria known as {\bf d-separation}. The d-separation criteria will become important not only in our discussion of causal discovery, but also in the identification of back-door and front-door adjustment criteria. Conceptually, d-separation is a measure of the potential association between a set of variables, and how to block that association.
    \\\\
    A path \(p\) is {\bf d-separated} by a set of nodes \(Z\) if and only if \cite{pearl2016causal_primer}:
    \begin{itemize}
        \item p contains a chain of nodes \(A \xleftarrow{} B \xrightarrow{} C\) or a fork \(A \xleftarrow{} B \xrightarrow{} C\) such that the middle node \(B\) is in \(Z\).
        \item p contains a collider \(A \xrightarrow{} B \xleftarrow{} C\) such that the collision node \(B\) is not in \(Z\), and no descendants of \(B\) is in \(Z\).
    \end{itemize}
    
    Now that we've briefly touched upon the fundamentals of causal graphical models, we now mention the first stage of causal analysis. The first stage of causal analysis is to specify the structure and orientation of the causal graph. This is typically either done by using domain knowledge to specify graph edges and directions, or domain knowledge coupled with what are known as causal discovery algorithms. Causal discovery algorithms fall loosely into the categories of constraint-based methods and score-based methods. Algorithms such as IC (Inductive Causation) and IC* (Inductive Causation with Latent Variables Such As Unobserved Confounders) \cite{Pearl:2009:CMR:1642718} rely on conditional independence constraints such as d-separation as their foundation, as well as the common graph sub-structures previously defined. Other well known algorithms such as PC \cite{PC_Orig} and FCI \cite{FCI_Orig} modify the exploration and evaluation procedure on edges to make causal discovery more practical for sparse cases. The most commonly known score-based method is referred to as Greedy Equivalence Search (GES) \cite{GES_Orig}. This method uses a Bayesian Information Criterion score based search method to identify the optimal causal graphical model. It is out of the scope of this paper to provide a full exposition on every algorithm. However, we briefly present the IC algorithm to provide the flavor of how constraint based methods work. Scoring based methods are a bit more complex, so see \cite{chickering2002learning} to get an understanding of how GES identifies equivalence classes through search and \cite{GES_Orig} to understand how this search takes place within equivalence classes.
    
    {\bf IC Algorithm} pulled directly from Judea Pearl's work \cite{Pearl:2009:CMR:1642718}:
    \begin{itemize}
        \item {\bf Input}: \(\hat{P}\), a stable distribution on a set \(V\) of variables.
        \item {\bf Output}: a pattern \(H(\hat{P})\) compatible with \(\hat{P}\). 
        \item {\bf Algorithm}:
            \begin{enumerate}
                \item For each pair of variables \(a\) and \(b\) in \(V\), search for a set \(S_{ab}\) such that \(a \perp b | S_{ab} \) holds in \(\hat{P}\). Construct an undirected graph \(G\) such that vertices \(a\) and \(b\) are connected with an edge if and only if no set \(S_{ab}\) can be found.
                \item For each pair of nonadjacent variables \(a\) and \(b\) with a common neighbor \(c\), check if \(c \in S_{ab}\). if it is, then continue. If is is not, then add arrow heads pointing at \(c\) from \(a\) and \(b\).
                \item In the partially directed graph that results, orient as many of the undirected edges as possible subject to: (i) Any alternative orientation would yield a new v-structure; or (ii) Any alternative orientation would yield a directed cycle.
            \end{enumerate}
    \end{itemize}
    
    A key point of note with causal discovery algorithms is that the graphs they output should always be checked. The methods are not developed to the point yet where they are so trustworthy that they do not warrant an in-depth evaluation by a domain expert. In practice, causal discovery is often paired with domain knowledge to define the final causal graph. A recent overview of causal discovery methods can be found in this great article written by Clark Glymour, Kun Zhang, and Peter Spirtes \cite{Causal_Discovery_Survey:10.3389/fgene.2019.00524} and in this practical guide \cite{Practical_CD:doi:10.1111/phc3.12470}.
    Causal discovery is an active area of research, with many new methods being developed \cite{NIPS2018_7930} \cite{Kernel_Based_CI} \cite{Monti2019CausalDW} \cite{ramsey2017million}.
    
    After the causal graphical structure has been defined, we can begin the tasks of causal identification and estimation. Again, a full and in-depth treatment of these topics is out of the scope of this paper. However, we briefly mention the core components of causal identification and estimation. 
    
    The first idea that's important to the estimation of causal effects is what is called 'do-calculus'. Do-calculus refers to a hypothetical, artificial manipulation of the cause and effect relationships in a causal graph. This manipulation is shown in the figures below. If the goal given this causal graph is to estimate the causal effect of \(X\) on \(Z\), then we would represent the pre-intervention outcome probability of \(Z\) as \(Pr(Z=z |X=x)\), and the post-intervention outcome probability of \(Z\) as \(Pr(Z=z | do(X=x)\). The pre-intervention distribution would be calculated from sub-figure a below, while the post-intervention calculation would be done using the causal graph form sub-figure b below. Thus, do-calculus ultimately amounts mathematically to taking the pre-intervention graph, and removing all inbound causal edges to the variable we want to estimate the effect for (in this case \(X\)). 
    
    \begin{figure}[htp]
        \centering
        \label{figur}\caption{Do-calculus as Graph Manipulations}
        \subfloat[Pre-Intervention]{\label{figur:1}\includegraphics[width=3cm, height=3cm]{fork_cgm.png}}
        \subfloat[Post-Intervention]{\label{figur:2}\includegraphics[width=3cm, height=3cm]{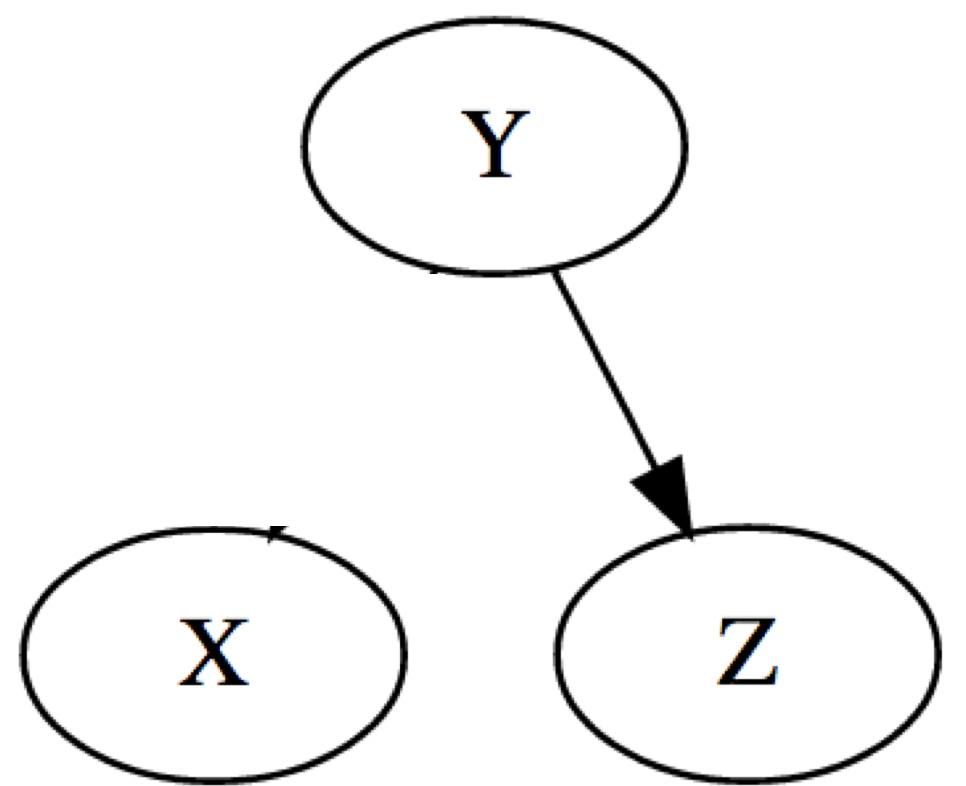}}
    \end{figure}
    
    The second idea that's important to the estimation of causal effects is how average treatment effect is defined for causal graphs. If we assume a variables \(Z\) and \(X\) are binary in the example causal graph, then the average treatment effect is defined as \(ATE = Pr(Z=1 | do(X=1)) - Pr(Z=1 | do(X=0))\). Notice that this looks very familiar in form to the definition of average treatment effect for the potential outcome framework. A more general definition of average treatment effect for causal graphical models over continuous treatments is \(d/dx(E[Pr(Z|do(x)])\).
    
    The third idea that's important to the estimation of causal effects is how to calculate probability statements defined using do-calculus. To compute the do-calculus statement \(Pr(Z=z | do(X=x)\), we introduce the backdoor and front-door criteria. The back-door criteria states:
    
    {\bf Backdoor Criteria}: Given an ordered pair of variables \((X,Y)\) in a directed acyclic graph G, a set of variables  \(Z\) satisfies the backdoor criterion relative to \((X,Y)\) if no node in \(Z\) is a descendant of \(X\), and \(Z\) blocks every path between \(X\) and \(Y\) that contains an arrow into \(X\).
    
    If a set of variables \(Y\) satisfies the backdoor criterion for \(X\) and \(Z\), then the causal effect of \(X\) on \(Z\) can be calculated as:
    
    \begin{center}
        \begin{equation} \label{eq2}
            Pr(Z=z | do(X=x)) = \sum_y Pr(Z=z | X=x, Y=y) Pr(Y=y)
        \end{equation}
    \end{center}
    
    This equation is called the backdoor adjustment formula, and can be used to calculate do-calculus based probability statements while using the pre-intervention causal graphical model specification, and the data. This estimation formula also puts no requirements on the particular method of estimating the probabilities, and can be used with any classical statistical estimation methods.
    
    A second situation that's important is when a causal estimation needs to be performed given there are known or potentially assumed unobserved confounders. This situation is illustrated in the figure below in which the variable \(U\) represents the unobserved confounding variable between \(X\) and \(Y\). Given the goal is to estimate the effect of \(X\) on \(Z\), and the variable \(Y\) is observed, then we can define a criteria called the front-door criteria, and can outline an adjustment formula for calculating \(Pr(Z=z | do(X=x))\) in this case.
    
    {\bf Front-door Criteria}: A set of variables \(Y\) is said to satisfy the front-door criterion relative to an ordered pair of variables \((X,Z)\) if
    \begin{enumerate}
        \item \(Y\) intercepts all directed paths from \(X\) to \(Z\).
        \item There is no unblocked path from \(X\) to \(Y\).
        \item All backdoor paths from \(Y\) to \(Z\) are blocked by \(X\).
    \end{enumerate}

    If a set of variables \(Y\) satisfies the front-door criterion relative to \((X,Z)\) and if \(Pr(X,Y)>0\), then the causal effect of \(X\) on \(Z\) is identifiable and is given by the formula:
    
    \begin{center}
        \begin{equation} \label{eq2}
            Pr(Z=z | do(X=x)) = \sum_y Pr(y|x) \sum_{x'} Pr(z| x', y) Pr(x')
        \end{equation}
    \end{center}
    
    \begin{figure}[htp]
        \centering
        \includegraphics[width=2.5cm, height=6.5cm]{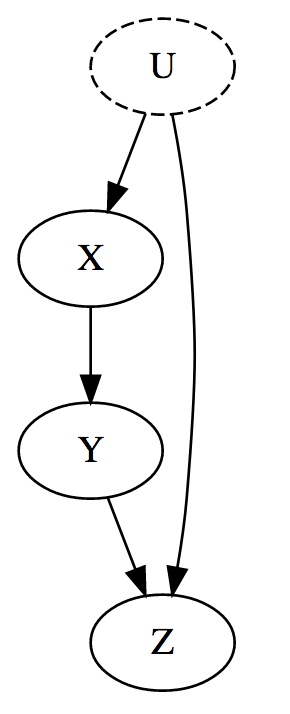}
        \caption{Front-Door Scenario With Unobserved Confounders}
    \end{figure}
    
    Using these adjustment criteria, along with the main setup of causal graphical models, a wide variety of causal estimations can be made given observational data. For a more in-depth treatment of the topic, we refer the reader to \cite{pearl2016causal_primer} and \cite{Pearl:2009:CMR:1642718} where our condensed exposition was gathered from. For the reader curious about how the potential outcome and causal graphical model frameworks are related, Judea Pearl provides an in-depth discussion of the connections here \cite{Pearl:2009:CMR:1642718}.

\subsection{Applications}

\subsubsection{Social Sciences and Medicine}

    Much of the application of causal inference appears in the domains of Social Science and Medicine. Studies and experiments performed in these domains typically have bias due to concerns of ethics, of non-control of environments, or of experimental units that are not comparable. For example, in the field of medicine, many times health related studies can only be performed in quasi-experimental or observational study setups because it would be unethical to perform otherwise. The classic example is for studying the effects of smoking. To produce a controlled random experiment, some non-smokers would have to be forced to smoke which is viewed as unethical. Even if this was the case, there are still potentially biasing issues due to sample selection bias.
    
    The social sciences have seen application of causal inference for the study of government policies on things like election results, levels of violence, poverty, etc. Rubins describes applications of causal inference in the social sciences at length in his book \cite{imbens_rubin_2015}. Other recent works on the role and application of causal inference in the social sciences include \cite{grimmer_2015} \cite{ogburn2017causal} \cite{Causal_D_Paper}.
    
    The medical domain has also seen a wide application of causal inference methods \cite{doi:10.2105/AJPH.2004.059204} \cite{doi:10.1080/00031305.2019.1575771} \cite{KLEINBERG20111102} \cite{imbens_rubin_2015}. While causal inference in the medical domain has traditionally been used for causal estimation in quasi-experiments and observational studies, it has also been applied to provide value in it's own right towards the discovery of things such as adverse drug reactions for the re purposing of existing drugs \cite{CAI20177} \cite{adv_dd}.

\subsubsection{Customer Experience}

    Customer Experience can be loosely defined as the experience a customer has when using a product or service, as well as when interacting with the company that produced the product or service. More and more companies are focusing on improving their customer experience as a differentiating factor in their industry. Some major aspects of a customer's experience include a customer's ease of setup and usage of a product, the ease with which self service documentation can be used to gain a return on investment from a product, post purchase customer engagements with a company, a company's support services, and a customer's ability to achieve their highest value possible while using a product. Most, if not all of these questions rely upon an understanding of customers, their behaviors, and their motivations for interacting with a product, service, or company resource. In this context, it's not enough to simply predict a customer's behavior. Being able to predict if a customer will log in to a product falls short of the goal of improving the customer experience. Companies need to understand why a customer logs in, what motivates them to purchase and use a product, what causes them to achieve success using the product, and what causes them to fall short of achieving success. Thus, data science for improving the customer experience is directly tied to not only causal inference, but to the big 3 questions in general. Some companies that are currently using causal analysis for improving the customer experience include Cisco, Microsoft, Uber, and Netflix,. 
    
    Cisco uses causal analysis for a wide variety of problems, including not only for improving the customer experience, but also for understanding the causes for renewals, the causes for customer behavior, for quantifying the effect of business activities on the business bottom line, and for developing causally driven policies for improving all of these areas. The application of causal analysis is critical to the success of Cisco's business because Cisco has a huge number of highly complex and diverse products, many of which interact to solve complicated problems. In this kind of environment, it's extremely hard to identify what's causing user behavior, what the motivations of a customer are, and what specifically is causing a customer to achieve value out of their product and service usage. Thus, Cisco applies all of the causal analysis methods mentioned here and more in an effort to improve the customer's experience.
    
    Microsoft is another company that has poured investment into the investigation of cause and effect for years. In fact, many of the core causal inference methods cited in this paper were published by researchers working at Microsoft. They've not only developed research in this area, but have also developed the DoWhy \cite{dowhy} and EconML \cite{econml} python libraries. DoWhy focuses on causal graphical model methods for direct estimation while EconML focuses on more recent and advanced machine learning and regression based methods for estimating cause and effect. 
    
    Uber has also recently published blogs on their general application of causal inference for improving the customer experience when using their product \cite{CI_Uber_harinen_li_2019}. They've also written on specific topics in causal inference for measuring the causal factors of their business bottom line such as mediation modeling \cite{MM_Uber_li_harinen_2019}. Uber has also recently released a python package for causal inference using uplift modeling \cite{uber_causalml}.
    
    Netflix is another company that seems to have began investigating the use of causal analysis. While they don't seem to have published any major results or work efforts around using causal analysis, they seem to have set up a research group for it \cite{mcfarland_pow_glick_2019}.
            
\section{Intelligent Action}

    In this section, we outline fields and methods related to acting intelligently in complex environments. Companies typically want to understand how effective the actions they take are on the metrics they care about (Eg. revenue, customer satisfaction, return on investment, recurring revenue, etc.). Furthermore, they want to be able to develop strategies for taking action given a wide variety of conditions in complex, changing environments with potentially many agents. To help answer these kinds of questions, we turn our attention to the wide variety of fields and methodologies for intelligently making decisions, optimizing the control of systems, and acting in complex environments. We tend to interchange the terms 'decision' and 'action' since a decision in the business context is usually a decision to take a particular action.
    
    We start with the area of decision theory and how this field lays the foundations for making decisions. We then move on to the idea of influence diagrams, and how these can be used to determine optimal decision making and actions (and are distantly related in spirit to causal graphical models). We then delve into reinforcement learning, a discipline related to the larger field of intelligent agents, and how it can be used to develop autonomous agents that can act in complex environments, and can also provide insights businesses insights into optimal action policies. Finally, we finish up with a discussion of game theory and some of the fundamentals of formulating intelligent action problems when there are multiple actors in the environment.
    
    In the applications section, we discuss how systems for intelligently acting are applied to electronic control systems, marketing environments, and to customer experience and engagement applications.

\subsection{Decision Theory}

    Decision theory is the study of how individuals and groups make or should make decisions \cite{choices_dt_1998}. In a world of risks, rewards, and uncertainty, there is a need to understand how to make decisions and take actions to reach our goals. This requires an understanding of both what decisions should be made from a theoretical standpoint, and how decisions are actually made from a realistic standpoint. These views of decision making are referred to as normative and descriptive theories of decision making respectively. Both views are potentially valuable from a business perspective. Given a formulated business goal or metric, a company wants to know what actions it can take to reach the defined goal or improve the metric. This requires normative decision theory to understand the actions a company should theoretically take to reach it's goals. However, descriptive theories can come into play depending on how the action is being executed, or in what environment. For example, a company may be trying to increase the sales of a specific product. However, whether or not to buy a product is up to the customer, not the company. Thus, while a company takes action to influence the customer, the customer is the one making the decision. In this case the company could be formulating it's own actions from a normative perspective, while modeling the customer's actions from a descriptive perspective. 
    
    Many challenges arise in decision making and decision theory. For example:
    \begin{itemize}
        \item \textbf{Modeling the Problem}: How should decision making be modeled in a precise way that can be analyzed? How should decisions be mathematically compared and ranked? How can decisions making be defined as 'optimal' in a mathematical way?
        \item \textbf{Ignorance}: How can decisions be made given a lack of information?
        \item \textbf{Uncertainty}: How can decisions be made given various sources of uncertainty? There could be uncertainty about the effect of an action, uncertainty about the reward of an action, and uncertainty about the state of the world.
        \item \textbf{Risk}: How can decisions be made given competing objectives? A company may want to take actions in a risk avoiding or risk accepting way.
    \end{itemize}
    
    The following subsections outline these challenges and the solutions that decision theory provides. Most of the work done in the area of intelligent agents, reinforcement learning, and game theory use concepts and methods from decision theory as a baseline. Decision theory is ultimately the guiding foundation for being able to understand and to prescribe decisions and actions in goal directed ways, and have a high amount of potential to improve business bottom line metrics. The aspects of decision theory we outline below are framed largely after the descriptions provided in Michael D Resnik's book 'Choices' \cite{choices_dt_1998}. 

\subsubsection{Modeling the Problem}

    Decision problems are generally modeled in terms of states, actions, next states, and rewards. In the problem of decision theory, an agent typically starts with some current state. From this current state, an agent has a set of choices or actions. These actions can be discrete or continuous, but to simplify analysis we assume that they are discrete. Given the current state, taking an action is assumed to produce a possible next state. What this next state is could be stochastic where actions only have a probability of producing a particular state. A utility value (reward) \(U(S')\) is assigned to the next state which describes how desirable this state is. This decision making can be framed in terms of what is known in decision theory as a decision tree. This decision tree is different from the decision trees that are typically used in prediction and pattern mining. The machine learning based decision trees are mathematical models for performing a specific task such as classification or regression. Decision trees in this context are models of an agent's possible interactions with an environment, and the likely results of those actions. The framing of decision making problems into states, actions, next states, and rewards is a simple yet powerful model. It is used in nearly all methods of understanding and learning to act in complex environments. The figure below provides an example decision tree. The choice of state and action representation is not generally restricted. Both states and actions can be discrete or continuous. The choice of utility function however is a bit more nuanced. We will not delve fully into the requirements of utility functions. However, we will mention that in most decision making scenarios, a utility function is required to adhere to a set of preference ordering rules (allowing an agent to rank one state over another or to be indifferent to them). Some decision making scenarios require further restrictions on utility functions, like they must not only contain ordering information but also interval information that quantifies the relative degree of one preference over another. This restriction is important for the decision making using expected utility (introduced in the next section) which underlies most decision making algorithms used in practice today.
    
    \begin{itemize}
        \item \textbf{States}: A numeric representation of all relevant components of the environment for decision making. It typically includes information about aspects of an environment, as well as information about the decision maker in an environment.
        \item \textbf{Actions}: A numeric representation of all possible choices available to a decision maker given a current state in the environment.
        \item \textbf{Rewards}: A numeric representation of the value of states, or the value of state-action pairs. These rewards are represented as utilities which map a decision maker's value scale and ordering of every state of state-action pair. 
    \end{itemize}
    
    The decision tree in figure 2 shows an case where there are two actions given a current state, each of which has a probability of producing a next state (represented by the chance nodes), along with a reward or utility. In the next major section on reinforcement learning, these possibilities are represented as state, action, reward, and next state tuples for learning to estimate the value of sequential decisions automatically.

    \begin{figure}[h]
        \begin{center}
            % Set the overall layout of the tree
            \tikzstyle{level 1}=[level distance=3.5cm, sibling distance=3.5cm]
            \tikzstyle{level 2}=[level distance=3.5cm, sibling distance=2cm]
            
            % Define styles for bags and leafs
            \tikzstyle{bag} = [text width=4em, text centered]
            \tikzstyle{end} = [circle, minimum width=3pt,fill, inner sep=0pt]
            
            % The sloped option gives rotated edge labels. Personally
            % I find sloped labels a bit difficult to read. Remove the sloped options
            % to get horizontal labels. 
            \begin{tikzpicture}[grow=right, sloped]
            \node[bag] {State}
                child {
                    node[bag] {Chance}        
                        child {
                            node[end, label=right:
                                {Reward 4}] {}
                            edge from parent
                            node[above] {Next State 2}
                        }
                        child {
                            node[end, label=right:
                                {Reward 3}] {}
                            edge from parent
                            node[above] {Next State 1}
                        }
                        edge from parent 
                        node[above] {Action 2}
                }
                child {
                    node[bag] {Chance}        
                    child {
                            node[end, label=right:
                                {Reward 2}] {}
                            edge from parent
                            node[above] {Next State 2}
                        }
                        child {
                            node[end, label=right:
                                {Reward 1}] {}
                            edge from parent
                            node[above] {Next State 1}
                        }
                    edge from parent         
                    node[above] {Action 1}
                };
            \end{tikzpicture}
            \caption{}
        \end{center}
    \end{figure}

\subsubsection{Ignorance}

    The idea of ignorance in decision making is different from the idea of uncertainty. Uncertainty applies to decision making when we have uncertainty about the information used for making decisions, the effect of decisions, and the value of decisions. Uncertainty is applied when information exists, but it is unclear how much the information must be trusted. Ignorance applies to decision making when the uncertainty is absolute and no information exists for evaluating decisions other than preference ordering assignments to state action combinations. Given the need to act under ignorance, and given a set of state action pairs that have been assigned a preference ordering, the following four rules for choosing optimal actions have been proposed.
    
    \begin{itemize}
        \item \textbf{Maximin Rule}: Given a set of action preferences over all possible states, choose the best of the worst. May as well assume pessimistically that the action and state combination are the worst for that action. So, get the worst utility for all actions, and pick the best out of the worst. The challenge with this rule is that is fails to take advantage of actions that have slight losses but large possible gains.
        
        \item \textbf{Minimax Regret Rule}: Regret is the amount of missed opportunity for a state act pair. The idea is to first calculate the regret of a missed opportunity for each action, and then pick the action with the minimum out of the maximum regrets. In other words, pick the action that has the lowest potential opportunity cost.
        
        \item \textbf{Optimissim-Pessimisim Rule}: Linearly scale between pessimistically assuming the worst utility for each action state pair will occur and optimistically assuming the best utility for each action state pair will occur.
        
        \item \textbf{Principle of Insufficient Reason}: The previous rules only considered maximums and minimums, and not the intermediate utility values, or the utility distributions. Thus, assign a uniform distribution to all states and find the expectation of the utility of an action over all states. 
    \end{itemize}
    
    Many other challenges and paradoxes can arise in the context of decision making while using preferences and utility as a basis for measuring the desirability of reaching a state given a particular action was taken. To resolve some of these issues, decision theorists have attempted to describe how a rational agent would act under specific conditions. The conditions are called the mixture condition and irrelevant expansion condition. The mixture condition indicates that, given two actions that an agent is indifferent to taking, it is then also indifferent to flipping a coin and taking the first action when heads and the second when tails. The irrelevant expansion condition indicates that, given an initial set of states and actions and an optimal action, the introduction of new actions should not change the initially selected action. These conditions eliminate all but the last rule which states that the action that maximizes expected utility should be the one chosen. In the later sections of influence diagrams and reinforcement learning, the idea of finding the action that maximizes the expected utility and expected returns. The next sub section discusses how uncertainty and thus probability enters into decision making.
    
\subsubsection{Uncertainty and Risk}

    Uncertainty exists in every aspect of the world and decision making. Will the sun rise tomorrow? We might be confident that the sun will rise tomorrow, but there still may be some small amount of uncertainty. Ultimately, probability and statistics were developed to help mathematically reason in a precise and measured way given a world of uncertainties. Statistics is mainly categorized into the Frequentist approach or the Bayesian approach, which connect back to the ideas of absolute and relative truth. Frequentist statistics deals with a world of absolute truths. Parameters of interest in the world have absolute values and randomness in sampled data sets is responsible for the uncertainty of the parameters. Given enough observations over time, the sampled data sets using to create an estimate of the parameter of interest will converge to the true parameter. Concisely, the parameters of interest in an analysis are fixed and randomness is due to the data. Bayesian statistics deals with a world of relative truths. Parameters of interest may have fixed values, but even if they do we can never fully know what they are. Estimates of these parameters can only be understood as beliefs about the value of the parameter. Concisely, the parameters of interest in an analysis are random and the data is fixed. 
    
    For decision analysis, probability is used to quantify the uncertainty about the parameters of a decision making problem. This might be the uncertainty of an action producing a particular state, or of a state-action pair producing a particular value. This uncertainty provides a different context for making decisions than that of making decisions under ignorance. An agent in this context has some information or methods to estimate the value of an action given a state. Thus, decision making in this context requires not only reasoning about the value of states given actions, but also reasoning given there is some uncertainty in our estimates. The following list outlines how to make decisions using expected utility.
    
    \begin{enumerate}
        \item Given a set of state action pairs labeled with preferences that map to an 'interval scale' (aka the preference ordering doesn't change under linear transforms of the preference mapping). For a current state \(S\), can be viewed as a table with columns being the possible next state values \(S'\), rows being the possible actions from the current state st, and entries being the utility. We assume here that a proper interval utility scale has already been created, and is already available. For more information, see chapter 16 of \cite{Russell:2009:AIM:1671238} and chapter 4 of \cite{choices_dt_1998}.
        \item Given a set of probabilities assigned to each state action pair, which is the probability that the desired state is achieved given an action.
        \(Pr(S' | S, a) \buildrel d \over = to Pr(U(S') | S, a)\)
        \item Expected Utility is \(E[U(a|S)] = \sum_{i}Pr(U(S_i')| S, a)*U(S_i')\). A more formal qualification and set of proofs for the Von Neuman-Morgenstern Utility Theory is given in chapter 4 of the very approachable book 'Choices' \cite{choices_dt_1998}.
        \item Decisions are chosen by: \( a^* = \underset{a}{argmax} E[U(a|s)] \)
    \end{enumerate}
    
    The use of Bayes theorem can also be used to determine the value of having additional information in making a decision problem. It can be used to provide upper bounds on how additional information might change the probabilities associated with states and actions, which may in turn change the decisions made. In this case, let's assume that a state is made up of random variables \(X1\), \(X2\), and \(X3\). However, lets assume that the values for X3 are not observed. In this case, if it is assumed that the result is not conditionally independent of \(X3\), the expected utility is effectively marginalized over all values of \(X3\), resulting in a decision making problem where \(X3\) doesn't come into play. However, the business may want to know how valuable knowing the value of variable \(X3\) would be to have. Perhaps the business could purchase the information from a third party, but the cost may be significant. Calculating what is called the 'value of perfect information' tells us the effect of having this additional information would be. Thus, in the first equation below, we describe the expected utility given our current information \(X1\) and \(X2\). The second equation describes the expected utility given the new evidence, which simply adds to the random variables that are conditioned on in the probability used to calculate the expectation. Yet this isn't enough, as the exact value of \(X3\) isn't known. So, we can calculate the weighted average over the expected utility given the new information over all possible values of the new information. Using this, we get the third equation which returns the numeric value of this information. To use this information in determining whether or not to gather new information, a cost can be assigned to gathering the information of \(X3\). If the value of perfect information is greater than the cost of gathering the value of a specific random variable, then one should theoretically decide to gather the new information. The assignment of cost can be tricky when the utilities are not monetary in nature. The value of perfect information is the difference in expected utilities of the best action given the extra information and given no extra information. Therefore, the cost of gathering the new information must also be specified in terms of preferences, or must be back calculated somehow into it's component parts that align with some human measured value such as money, time, effort, etc.
    
    \begin{center}
        \begin{equation} \label{eq5}
            E[U(a|X_1, X_2)] = \underset{a}{max} \sum_{s}Pr(S' = s|a, X_1, X_2)U(S'=s) 
        \end{equation}
        \begin{equation} \label{eq6}
            E[U(a|X_1, X_2, X_3)] = \underset{a}{max} \sum_{s}Pr(S' = s|a, X_1, X_2, X_3)U(S'=s) 
        \end{equation}
        \begin{equation} \label{eq7}
            VPI_e(X_3) = (\sum_{k}Pr(X_3 = x_{k})E[U(a| X_1, X_2, X_3)]) - E[U(a|X_1, X_2)]
        \end{equation}
    \end{center}
    
    While expected utility theory is widely used in applications for decision problems and optimal decision making, some decision theorists have pointed out some potential issues. The more notable criticisms and paradoxes include Allais, Ellsberg, St Petersburg, and Predictor paradoxes. We forgo a deep discussion of these paradoxes \cite{choices_dt_1998}. We will however take note of the Predictor paradox, which introduces problems in which conditional probabilities are assigned to states given actions and historical information, where it is also known that these same states are not materialized because an action was taken. In this case, the best action suggested by the expected utility theorem differs from the best action suggested by the dominance principle (when one action's resulting state utility values are better than all other action's corresponding utility values). The issue arises because the conditional probability assignments of the a state occurring because an action was taken imply a causal link. To resolve the Predictor paradox, we will take the stance that these probabilities must be representations of the propensity for an action to bring about a state, making the causal linkage between state and act explicit. Proponents of what's called evidential decision theory suggest that the implicit causal linkage can be removed if decision theory is viewed as simply a method for understanding the preferences of results. We set that view aside and take the causal view because our primary goal is to understand how data science can be used to choose actions that will cause a desired business result to occur that may not have occurred otherwise. In the next section, this causal view is made more formal as a step for building influence diagrams where the goal is to create directed causal links between random variables, actions, outcomes, and utilities.

\subsubsection{Challenges}

    Many challenges arise in the field of decision theory. The first is that traditional decision theory quickly becomes untenable for complex environments or decision problems with multiple states and actions. Attempting to capture every possible set of states, actions, and rewards in a compact way using hand developed models is untenable. In later subsections, we outline methods such as intelligent agents and reinforcement learning that seek to handle the challenge of complex and large environments and actions in an automated way, without formal models of the environment they are working in. Another challenge is that decision modeling tends to take a long time to implement. A good example is when making a decision when driving. A driver would not do a decision analysis to determine every minute action they take. The analysis would take too long, and would not provide enough benefit to justify performing. However, when buying a house, a decision analysis may be warranted because we have enough time to perform it, and there is a large amount of benefit. From a business perspective, decision theory and decision analysis provide a good framework for attempting to construct decision making strategies with limited variables and actions, and the cost of the time investment for performing the analysis is out weighed by the value of the return or the risk of making a wrong decisions. Finally, there are many mathematically counter intuitive and philosophically paradoxical scenarios that still exist in decision theory that are still areas of research.

\subsection{Influence Diagrams}

    Influence diagrams (also known as decision networks) utilize ideas from decision theory, Bayesian networks, and causal inference to more formally outline the dynamics of variables in an environment, the causal relationships between these variables, and the causal relationship between an action and these environment variables and outcomes. Influence diagrams provide a mathematically structured way to understand various forms of causal relationships in an environment, and to prescribe actions to take to maximize the chance of achieving particular goal. In the following sub sections, we outline the principles of causality applied to actions and how influence diagrams are formulated. We frame largely in the same way as chapter 16 from this reference \cite{Russell:2009:AIM:1671238}.

\subsubsection{Causality in the Action Context}

    In the previous sections, we outlined what causal inference is and how causal inference can be applied to understanding random variables that characterize the data set. Causality comes into a business problem in two ways. The first is to understand how changes in one or more random variables affect another. This kind of causal analysis is performed to understand the dynamics of a problem or an environment. This is the second question of the big three, why is something happening? The second kind of causality comes into play when some agent or decision maker wants to interact with the environment and change the dynamics so that the eventual outcome will be more favorable. This is the second question of the big three, what is the optimal action to take that will cause the best chance for me to achieve my goals? Influence diagrams provide a method to incorporate both kinds of causal information so that multi-chain cause and effect relationships can be analysed. An example of this kind of multi-chain interaction is in sales revenue. A business may want to increase their amount of sales. A business may attempt to influence a customer from by sending emails to them. The content of these emails in turn may influence a customer to open an email. This, in turn, may cause them to purchase a product. There is a cause and effect relationship between a customer opening an email and them buying a product, as well as a cause and effect relationship between choices of content in an email and opening an email. Thus, understanding how much a choice of email impacts sales requires this multi-chain causal analysis. A diagram of this scenario has been provided in figure 2 below.

    \begin{figure}[h]
        \begin{center}
            \includegraphics[width=3.5cm, height=4cm]{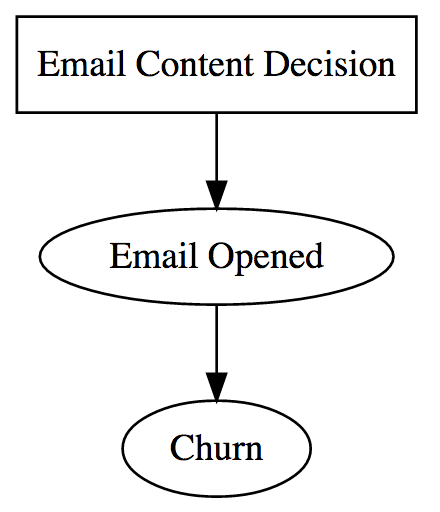}
            \caption{}
        \end{center}
    \end{figure}

\subsubsection{Extensions to Bayes Nets}

    In the 1980's, a new method of formalizing statistical relationships between variables was created \cite{pearl1985bayesian}. Judea pearl was credited with the accomplishment of creating Bayesian networks when receiving the A.M. Turing award in 2011. We won't go through the full idea of all aspects of Bayesian Networks. We instead introduce the core mathematical concepts, and how the method is extended to form influence diagrams. For a full tutorial on Bayesian networks, the following references are great introductions to the construction, learning, and inference with Bayesian networks \cite{Russell:2009:AIM:1671238} \cite{pearl1985bayesian} \cite{Murphy:2012:MLP:2380985}.

    Bayesian networks are a structured representation for a joint probability distribution that can be factored into conditional probability formulas. Specifically, the probability of any node \(X_i\) is conditionally independent of all other nodes given it's parents.

    \begin{center}
        \begin{equation} \label{eq1}
            \begin{split}
                Pr(X_n, ..., X_1) & =  Pr(X_n|X_{n-1}, ..., X_1)*Pr(X_{n-1},X_{n-2}, ..., X_1) \\
                & = Pr(X_n|X_{n-1}, ..., X_1)*Pr(X_{n-1}|X_{n-2}, ..., X_1)*...*Pr(X_2|X_1)*Pr(X_1) \\
                & = \prod_{k=1}^{n}Pr(X_k|\bigcap_{j=1}^{k-1}X_j)
            \end{split}
        \end{equation}
        
        \begin{equation} \label{eq2}
            Pr(X1, ..., X_n) = \prod_{i=1}^{n}Pr(X_i|parents(X_i))
        \end{equation}
    \end{center}

    \begin{figure}%
        \centering
        \subfloat[General Bayes Net]{{\includegraphics[width=2.6cm, height=5cm]{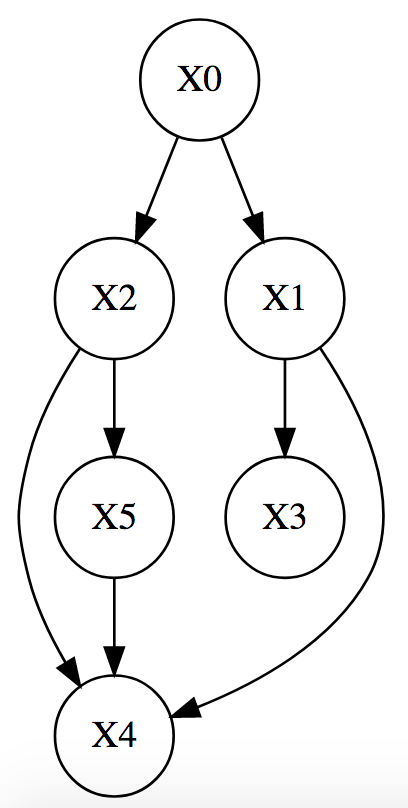} }}%
        \qquad
        \subfloat[Churn Risk Bayes Net]{{\includegraphics[width=8cm, height=5cm] {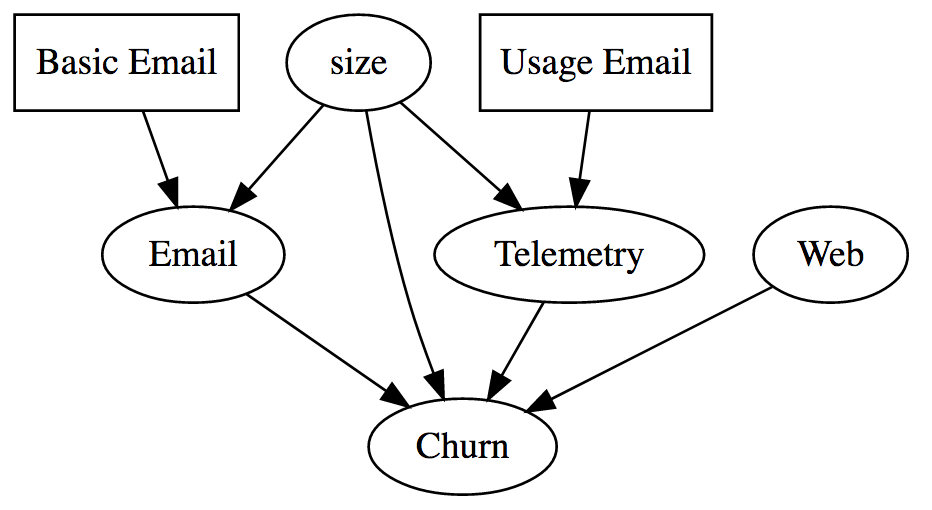} }}%
        \caption{2 Figures side by side}%
        \label{fig:example}%
    \end{figure}
    
    The major challenges in Bayesian nets include parameter learning, structure learning, and inference. Structure learning is the problem of identifying the links between random variables identified in a network. These links may be directed or undirected depending on the kind of net and the assumptions made by the algorithm. Structure learning is typically framed as an intelligent search problem where potential candidate network structures are proposed, evaluated, and then modified in an attempt to make the new network more reasonable than the current network structure. Parameter learning is the challenge of estimating the parameters of the conditional probability distributions for each network node. Common statistical optimization methods like expectation maximization are used to estimate the parameters of the conditional distributions at network nodes. Inference of unobserved variables refers to the challenge of taking known values of random variables (evidence) from a data sample, and using the network to estimate the value of random variables for which data was not observed. There exist exact methods of inference such as variable elimination, as well as approximate methods such as Markov Chain Monte Carlo methods, Variational Inference methods, Loopy Belief Propagation, and many others \cite{Murphy:2012:MLP:2380985}.
    
    Influence diagrams, also known as Decision Networks extend the structure of Bayesian networks to include complex interactions of random variables in the process of decision making. We draw on the basic descriptions provided in \cite{Russell:2009:AIM:1671238} here. Given information about the random variables defining a current state, the decision network represents an agent's possible actions, the resulting state given an action, and the utility of this next state. It does so by defining the relationships between chance nodes (represented in diagrams with ovals), decision nodes (represented in diagrams as rectangles), and utility nodes (represented in diagrams with diamonds) \cite{influence_diagrams_1984}. Chance nodes represent random variables, just as in bayes nets. The chance nodes represent both the current state and next state random variables in a decision problem. Decision nodes represent the actions or decisions available to influence an outcome that are available to a decision maker. Utility nodes represent an agent or decision makers utility function. The utility node's parents are all variables that directly effect the utility value. The utility node is accompanied by the exact calculation of the utility value for a given set of assignments to the utility node's parent nodes.
    
    To evaluate a decision network or influence diagram, perform the following:
    \begin{enumerate}
        \item Set the values of the random variables that define the current state.
        \item For each possible value of the decision node:
        \begin{enumerate}
            \item Set the decision node to that value.
            \item Calculate the posterior probabilities from the parent nodes of the utility node using bayes net inference algorithms.
            \item Calculate the resulting utility for the action.
        \end{enumerate}
        \item Return the action with the highest utility value.
    \end{enumerate}
    
    Thus, the influence diagram literally calculates the hypothetical utility outcomes given a decision or action. Since the random variable links in the bayes net are being used to make calculations in hypothetical environments, it is crucial that the links between random variables be causal in nature. If the links are not verified to be causal, then the hypothetical next states are not statistically valid from a decision making perspective. Constructing 'hypothetical' scenarios requires causal links. This is why taking a predictive model and varying the value of parameters identified to be important is not a valid way to explain the cause of model prediction values. 

\subsubsection{Steps to Modeling Decisions}
    
    Stuart Russell and Peter Norvig \cite{Russell:2009:AIM:1671238} suggest the following process for constructing and evaluating decision networks.

    \begin{enumerate}
        \item \textbf{Model the Problem}: Determine the random variables defining state and next state, determine the decisions possible.
        \item \textbf{Create a Causal Model}: Determine the causal structure of the model. This may be informed by experts, intuition, and ideally validated using the methods introduced in the last major section on causal inference.
        \item \textbf{Assign probabilities}: Assign causal probabilities using historical data sets and causal inference.
        \item \textbf{Assign utilities}: Assign utilities that align with the expected utility theorem to the possible outcomes. Given many possible outcomes, methods from multi-attribute utility theory can be used.
        \item \textbf{Verify and refine the model}: Develop and use a set of 'gold standard' data to validate the system against.
        \item \textbf{Perform sensitivity analysis}: Systematically vary random variable probabilities and utilities to understand how sensitive the model's prescribed decision is.
    \end{enumerate}
    
    The figure below corresponds to a decision diagram related to a business use case of deciding to send either a basic email, a usage email, or both and how these have a causal impact on churn or renewal risk. The diagram shows that the choices of email content affect the rate of email opens (the 'Email' node) and the amount of product usage seen in telemetry (the 'Telemetry' node). These in turn have a causal effect on churn, which in turn has an associated utility.

    \begin{figure}[h]
        \begin{center}
            \includegraphics[width=5.3cm, height=6cm]{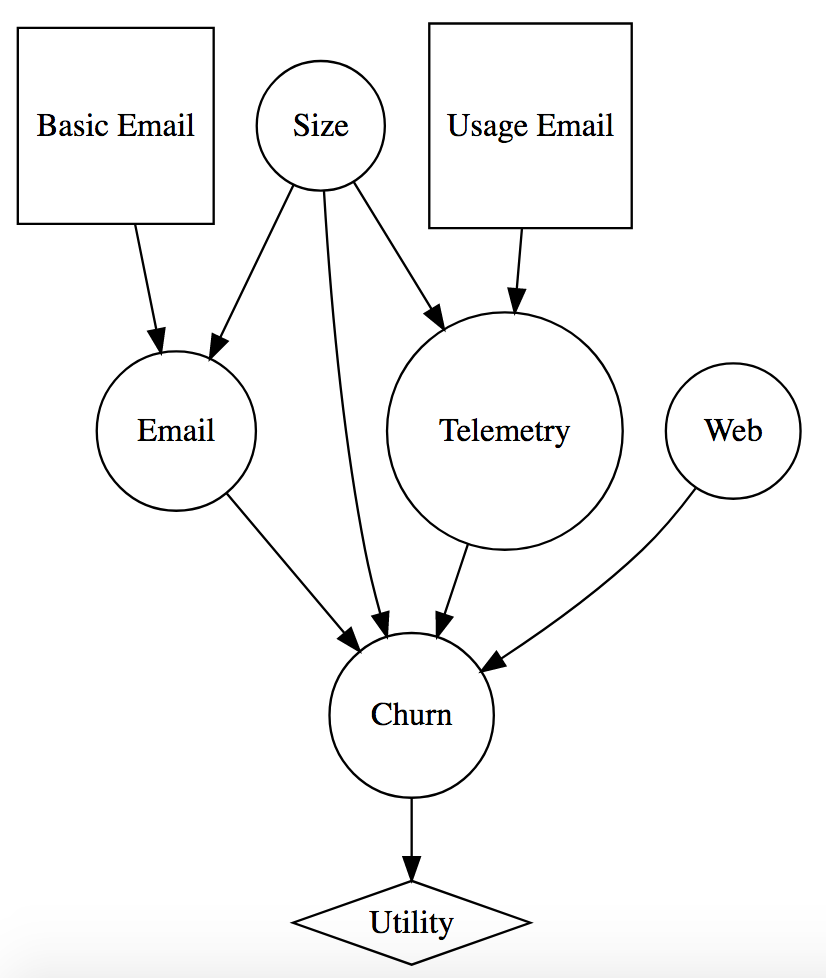}
            \caption{}
        \end{center}
    \end{figure}
    
    Dynamic decision networks extend the idea of decision networks to multi-stage decision making. They offer a methodology for performing and evaluating sequential decision making tasks by framing the problem as a partially observable markov decision process. Dynamic decision networks can be applied in the following way (as described in section 17.4.3 of this reference \cite{Russell:2009:AIM:1671238}).
    
    \begin{enumerate}
        \item Extend bayes nets by modeling the problem with dynamic bayes nets, which model states as well as their transition probabilities.
        \item Further extend the dynamic bayes nets with the addition of decision and utility nodes.
        \item Perform inference through the use of a filtering algorithm to estimate belief states given actions and observed values of random variables.
        \item Perform decision making by performing a bounded forward search of the state-action space to determine the best actions.
    \end{enumerate}

    Dynamic decision networks provide a method for learning how to act in complex environments as a natural extension of basic decision networks. However, this structured methodology can be computationally complex to evaluate not only due to the complexity of inference in bayes nets, but also because their use requires a forward search through state-action space. In the next major section, reinforcement learning is introduced as a alternative methodology for understanding and making decisions in sequence. Reinforcement learning frames the problem in a very similar way to dynamic decision networks. However, the practical application of reinforcement learning approaches the problem using approximate modeling methods that can generalize to new state action pairs, and are generally less computationally complex to learn.
    
\subsubsection{Challenges}
    
    Influence diagrams (Decision networks) and Dynamic Decision Networks have a few limitations. They typically require a deep understanding of the environment and the relationships between random variables, decisions, and utilities. The specification of utilities can also be complex to define, especially because utilities are required to conform to the expected utility theorem, which provides mathematical restrictions on how utility values are assigned to outcomes. Other challenges include the modeling limitations of bayesian networks, and the computational challenges that come along with performing inference with any bayesian network. The advent of reinforcement learning provides a mechanism to prescribe actions automatically that are significantly less computationally complex, can handle unspecified environment dynamics, and can generalize across large state and action spaces.

\subsection{Reinforcement Learning}

    Reinforcement learning is a long studied field in machine learning and artificial intelligence that pulls from the ideas of game theory, statistical decision theory, control systems, biological systems, and many others for learning to act in highly complex and changing environments. Reinforcement learning represents an advanced culmination of mathematical methods for allowing generally defined agents to learn to act optimally when environments are too complex to be able to specify the dynamics. The seminal book on reinforcement learning written by Richard Sutton and Andrew Barto describes reinforcement learning as "a computational approach to understanding and automating goal-directed learning and decision making" \cite{Sutton1998}. By making specific assumptions about the structure of the decision problem, specifying a reward function (which defines when something 'good' happens), and recursively defining the optimization criteria, we can come up with a computational method for learning to act in complex environments. The real value however is when we relax the assumption that we can fully specify the dynamics of the environment, and can instead learn to model the dynamics implicitly by learning 'action value functions' through intelligent trial and error. In the rest of this section, we present the concepts of reinforcement learning in a summarized form borrowed from the wonderfully approachable book written by Richard Sutton and Andrew Barto \cite{Sutton1998}. For a much more comprehensive foundation in reinforcement learning, reference their book.
    
\subsubsection{Framing the problem}

    The first component to understand about reinforcement learning is what mathematical objects exist in the decision making problem. Reinforcement learning consists of an agent and an environment. An environment is assumed to be everything outside of the control of the agent. The agent keeps track of the 'state' (all perceivable information about the environment relevant to making decisions) of the environment and decides to take a specific 'action' in an environment given its state, by using a 'policy'. The environment provides feedback or 'rewards' that indicate how well the agent is acting in the environment, as well as information that the agent needs to determine its state. Often, the process determining states and receiving rewards is described from the perspective of the agent, where the agent 'perceives' this information. Thus, the fundamental mathematical objects of the problem are 'states', 'actions', 'rewards', and a 'policy'. The general goal is to learn a policy that prescribes an action, which maximizes the rewards, given a state. The final mathematical object is called an 'episode'. An episode is a sequence of state, action, reward tuples which correspond to the history of actions taken by the agent, given a state, and the immediate reward that was returned. Episodes are of the form specified in equation 24 below \cite{Sutton1998}.
    
    \begin{center}
        \begin{equation}
            S_0, A_0, R_1, S_1, A_1, R_2, ...
        \end{equation}
        \begin{equation}
            Pr(S_t = s', R_t = r | S_{t-1} = s, A_{t-1} = a )
        \end{equation}
    \end{center}
    
    The second component to understand about reinforcement learning is how the decision making problem is specified. The decision problem for reinforcement learning is specified in terms of a 'Markov Decision Process'. A markov decision process defines the probability of the next state of the action and the immediate next reward given the agent's current state and that an agent takes a specific action from that state. A markov decision process is fully specified according to the following formulas (we adopt the specification made in the Sutton and Barto book). Note that we include the policy probability distribution for clarity and use later on, but it is not necessary to fully specify the structure of the decision making problem (aka the markov decision process).
    
    \begin{center}
        \begin{equation} \label{eq12}
            Pr(S_t = s', R_t = r | S_{t-1} = s, A_{t-1} = a )
        \end{equation}
        \begin{equation} \label{eq13}
            \pi(a|s) = Pr(A_{t-1} = a | S_{t-1} = s)
        \end{equation}
    \end{center}
    
    One of the key open questions in reinforcement learning is how to specify the 'reward function'. To eventually optimize behavior (aka maximize reward values), we have to map a reward event or signal from the environment (for example, winning a game of chess is an event that should be rewarded) to some numeric value. Usually heuristics and common practices are used to define reward functions which specify optimal behavior. For example, in an environment such as a game, a common way to frame the problem is to provide a value of +1 when the agent wins a game, -1 when the agent looses a game, and 0 for every other state-action pair. Even with reward functions specified as simply as this, reinforcement learning agents can typically learn to perform very well over time.
    
    The third component of reinforcement learning is the formal specification of the optimization problem. In classical reinforcement learning, optimal behavior is defined as a policy for choosing an action, given the current state of an agent, that maximizes 'total future cumulative rewards'. We've previously introduced the idea of a 'reward' signal, but this reward signal is immediate. This immediate reward is provided directly after an action is taken from a state. Given an immediate reward signal, the cumulative reward from time \(t\) is defined as the following:
    
    \begin{center}
        \begin{equation} \label{eq14}
            G_t = R_{t} + R_{t+1} + R_{t+2} + R_{t+3} + ... + R_T
        \end{equation}
    \end{center}
    
    Given this, we can define the 'value' of picking a particular action given a state as the following: 
    
    \begin{center}
        \begin{align} \label{eq15}
            v_{\pi}(s) & = E_{\pi}[G_t | S_t = s] \\
            & = E_{\pi}[R_t + G_{t+1} | S_t = s] \\
            & = \sum_{a}\pi(a|s)\sum_{s'}Pr(s', r | s, a)[r + v_{\pi}(s')]
        \end{align}
    \end{center}
    \begin{center}
        \begin{align} \label{eq18}
            q_{\pi}(s, a) & = E_{\pi}[G_t | S_t = s, A_t = a] \\
            & = E_{\pi}[\sum_{k=0}^{\infty}R_{t+k+1} | S_t, A_t = a ]
        \end{align}
    \end{center}

    \(v_{\pi}(s)\) specifies the value of a particular state. Specifically, it is the expected sum of future rewards from the current state. This is expanded using the definition of the reward function in equation 28. \(q_{\pi}(s,a)\) in equation 32 specifies the value of a particular combination of state and action. The value of \(q_{\pi}(s,a)\) the statistical expectation given the policy probability distribution and cumulative reward (\(G_t\) is cumulative reward) given the state s and action a. This fully quantifies the optimization goal of reinforcement learning.
    
\subsubsection{Bellman Optimality Equations}
    
    The fourth component of reinforcement learning to understand is how all of this comes together to specify an algorithm for learning to act optimally. The key to optimization in this problem is to realize that the current total cumulative reward Gt can be re-written recursively. Given this simple change of the optimization formulation, we can use iterative methods to learn a policy that optimizes our function \(q_{\pi}(s,a)\). 
    
    \begin{center}
        \begin{align} \label{eq20}
            q_{*}(s, a) & = \underset{a}{max} E[R_{t+1} + \underset{a'}{max}q_{*}(S_{t+1}, a') | S_t, A_t = a] \\
            & = \sum_{s', r}Pr(s', r | s, a)[r + \underset{a'}{max}q_{*}(s', a')]
        \end{align}
    \end{center}
    
    This equation can be optimized in a two step process. The general idea is to fix the policy probability distribution \(\pi(a|s)\) for choosing an action given a state. Given this is fixed, we optimize the value function, \(q_{\pi}(s,a)\), of a particular state-action pair. This is called 'policy evaluation'. Once the value function \(q_{\pi}(s,a)\) has been optimized, we fix it's values, and then update the policy distribution. To do this, we iterate over all state-action pairs and adjust the policy probability distribution to favor the action with the highest \(q_{\pi}(s,a)\) value for a given state. This is called 'policy improvement'. By alternating between these two methods, we can converge to an optimal policy. For more details, see the Sutton and Barto book \cite{Sutton1998}. Together, this method of alternating defines a general optimization process referred to as 'Generalized Policy Iteration'. This method forms the foundation of how nearly all reinforcement agents learn to act optimally in an environment. Furthermore, by inspecting the \(q_{\pi}(s,a)\) function, we can gain valuable insights into the relative value of individual actions. In the applications section, we describe how this knowledge can be used to strategically approaching the design and improvement of customer experiences.
    
\subsubsection{Approximate Model Free Methods}
    
    The fifth key idea in reinforcement learning is the idea of 'model free' methods. Model free reinforcement learning is performed when an agent does not have a model of the environment dynamics. Specifically, we do not know the function:
    
    \begin{center}
        \begin{align} \label{eq22}
            Pr(s', r | s, a)
        \end{align}
    \end{center}
    
    In this case, we have to estimate the values of \(q_{\pi}(s,a)\) using historical experience. We leave the in depth details of how to do this to the Sutton and Barto book's chapters 5 and 6. This important relaxation allows us to apply reinforcement to a wide variety of problems, and is one of the reasons that reinforcement learning is such a general and powerful method for answering questions related to intelligent actions.
    
    \begin{figure}[htp]
        \centering
        \includegraphics[width=11cm, height=4cm]{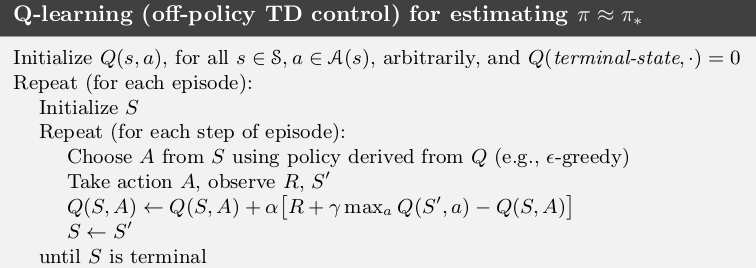}
        \caption{q-learning algorithm as defined by Sutton and Barto \cite{Sutton1998}.}
    \end{figure}
    
    The sixth and final concept in reinforcement learning we mention is the use of machine learning predictive models to represent the function \(q_{\pi}(s,a)\). Much of the hype around deep reinforcement learning revolves around how to apply deep neural networks as the machine learning model for estimating the value function effectively \cite{mnih2015humanlevel}. We leave the introduction of using machine learning methods for estimating this function to the Sutton and Barto book's chapters 9 through 11. The main idea is that with the use of machine learning, we can generalize over very large and even continuous state spaces, allowing us to optimize actions being taken in extremely complex environments. Reinforcement learning represents one of the most advanced and most general methods for determining how to optimize actions in the current field of AI and Machine Learning.

\subsubsection{Challenges}

    While reinforcement learning has gained wide application in simulated decision making environments, it still has challenges in real world application \cite{Thrun1993IssuesIU} \cite{real_world_rl_challenges}. One major criticism of reinforcement learning is that it can be very sensitive to the choice of hyper-parameters and model parameters used for approximating the value functions. In complex environments agents can take millions of trials to learn even a policy that performs mildly well, and in some cases fails to do that if the approximation model is ill-specified. Due to this issue, reinforcement learning research and application is still primarily done using simulations of the real world environments where agents can act in quick running simulated environments, allowing designers to tune agent hyper-parameters and model parameters. Other challenges include learning in risk averse environments (like self driving cars) where an agent still needs to learn about the results of potentially dangerous environments, the long burn in time where agents must spend large amounts of time initially learning the results of their actions, the ability to transfer knowledge learned to alternative environments, the right schedule for exploring versus exploiting a learned policy, and the ability to learn from small data sets efficiently. These areas are being actively researched today, and will go a long way towards making reinforcement learning more robust for industry applications. 

\subsection{Game Theory and Multiple Agents}

    Agents are almost never operating in an environment by themselves. Agents are simultaneously competing and collaborating with other agents and entities to meet a variety of goals. The study of acting intelligently in a shared environment with and against other agents that have shared and conflicting goals is called game theory. Companies, organizations, customers, and many others can all be thought of as agents operating in a shared environment with shared and conflicting goals. In this section, we aim to briefly outline the main aspects game theory, and discuss how to frame problems in a way that allows us to algorithmically answer how to act intelligently in this kind of environment. For more extensive discussions of the topic, we point the reader to the following resources \cite{Bonanno2018-BONGT-2} \cite{spaniel2011game} \cite{tadelis2013game}.

\subsubsection{Games}

    Game theory represents the natural evolution and extension of decision theory. While decision theory seeks to understand optimal decision making in an environment where only the decision makers decisions must be evaluated, game theory seeks to understand the optimal decision making strategy when there are multiple decision makers with shared or conflicting goals. This is of prime importance to the development of algorithms that can prescribe intelligent actions because decisions are rarely made in and isolated environment with one decision maker. Games are framed in an analogous way to the original decision making problem framing. The main components of games in game theory are,
    \begin{itemize}
        \item \textbf{Multiple Agents}: A game has multiple agents, each seeking to take actions that lead to states with maximal utility.
        \item \textbf{Actions}: Each agent has it's own, unique set of actions that can be taken given the state of the agent and the environment. These actions are also typically restricted given a well defined set of rules that are associated with the overall game.  These actions may also be informed through communication with other agents.
        \item \textbf{States}: States in game theory typically have a shared and a non-shared component depending on the kind of game. The shared component is typically a set of number representing aspects of the environment fully or partially visible to all agents. The non-shared component of a state may be the particular context of an individual agent within an environment described by a state. Depending on the visibility of an agent, the context of the other agents may be known, and factored into the decision making process. Other times, this is not known which adds to the complexity of reasoning about decision making.
        \item \textbf{Rewards}: Rewards in game theory are similar to those of decision theory. They still represent scale based utility functions. However, these functions can become much more complicated when the utility of states for a specific agent factor in not only the utility for that agent, but also for the set of other agents interacting with the environment. Sometimes utility values of a specific agent's state are directly related to the utility values of one or more opposing or collaborating agents. 
    \end{itemize}
    
    Game theory also makes various assumptions about the games and agents acting in a game to simplify the mathematical reasoning and modeling. Most of these assumptions follow logically from the definitions of rational behavior of goal oriented agents in general. The main set of assumptions are,
    \begin{itemize}
        \item Assumes games have utility scales associated with their outcomes.
        \item Assumes a given choice of strategies by opponents, each player chooses the strategy to optimize expected utility.
        \item Any two games with the same game trees are considered the same games.
    \end{itemize}

    Given this general description of the problem, game theory concerns itself mainly with two challenges. The first is the challenge of designing optimal agents. Here the goal is to understand how to develop intelligent agents that can learn strategies to maximize their expected utility throughout a game. The second challenge is called mechanism design. Here the goal is to develop games such that, given the game and the rules of the environment, the collective good of all agent's will be maximized by each agent maximizing their own utility. Given our interest in understanding how to prescribe optional actions in an environment, we focus on the first challenge. 

    Given the goal of acting optimally in games, game theory further breaks down games into a wide variety of types. Given this breakdown, the goal is to determine ways to construct an optimal policy for acting. Some of the ways to categorize games include,
    \begin{itemize}
        \item \textbf{Number of Agents}: Games with variable numbers of agents are typically broken down into two player games, and n-player games. Most of the classical analysis of games has been performed under the view of two player games. While simplifying analysis and providing insights into the basic challenges that arise in decision making due to the introduction of a single new agent, two player game analysis is limited. N-player games while general are much more challenging to analyze than two player games because of the exponentially more complex dynamics at play. 
        \item \textbf{Visibility}: Games in which the environment is fully visible are referred to as 'perfect information' games. Games in which the environment is only partially observable are referred to as 'imperfect information' games.
        \item \textbf{Chance}: Games which have stochastic elements are referred to as games with chance.
        \item \textbf{Competitive/Collaborative}: Games in which agents are all competing with each other and have conflicting goals are called 'strictly competitive' games, and games that allow for shared goals are called 'non-strictly competitive' games.
    \end{itemize}
    
    Other more specific categorizations exist such as static, dynamic, zero-sum, all exist. For a more comprehensive taxonomy of methods, we refer the reader to \cite{game_types_taxonomy}. A full treatment of all aspects of game theory are beyond the scope of this paper. However, we will mention a few central ideas from game theory that apply to acting intelligently in complex, multi-agent environments.
    
    The first central idea is that finding a policy to act intelligently can still be viewed as a form of search problem. Given the space of all possible policies, the problem is to efficiently search for the policy that will maximize expected utility, conditioned on the predicted behavior of all other agents and the dynamics of the environment. Therefore, some of the methods for solving certain games use optimization and modified search algorithms. One key algorithm to solve simple two person games is the 'minimax' search algorithm. This search algorithm suggests that the action an agent should take in the current state should be the one that minimizes the maximum expected utility of the opponents optimal action from the next state. We defer the entire implementation here \cite{Russell:2009:AIM:1671238}, and simply specify that this is an example of a search algorithm that is used to determine an agent's optimal policy. 
    
    The second major idea is that of equilibrium. An equilibrium is a condition under which each agent is assigned a policy such that no agent can benefit by switching its policy. Equilibrium are effectively local optimum where the dynamics of the environment due to the agents is fixed because no change in agent policies can cause an increase in utility. John Nash proved that every game has at least one equilibrium, which is typically called the Nash equilibrium. Equilibrium are important because most game theorists agree that a solution must necessarily be a Nash equilibrium. A closely related idea is Pareto Optimality. An outcome to a game is considered Pareto Optimal if there is no other outcome that all players would prefer. The idea of optimality and equilibrium are often combined when attempting to evaluate proposed solutions to specific kinds of games.
    
    Another major idea is the idea of n-player games and the evolving complexity of decision making with the addition of more agents. The representation and modeling of the dynamics of complex games (particularly N-player games) is extremely challenging, and partially related to the study of Complex Systems. Complex systems are systems which are composed of fully or partially independent phenomena that are all collectively interacting with each other. Complex systems include finite automota, biological systems, multi-agent games, and many other kinds of environments \cite{complexity_sciences_map}. The field draws upon the mathematics from a wide variety of other fields that seek to model highly non-linear systems composed of multiple smaller components. We mention field of complex systems as a promising area for solving problems within game theory, and for understanding how to act intelligently in realistically complex environments. 
    
    Finally, we mention the growing research field of multi-agent reinforcement learning (MARL). MARL is still in its early research stages, but algorithms developed have already been applied to games and game theoretic environments. While lots of research still needs to be done to make MARL agents ready to be applied to industry problems, they offer a promising advancement in the understanding and modeling of intelligently acting in realistic environments.

    There are many other challenges and ideas in game theory that we haven't discussed here. The field of game theory is of crucial importance to the goal of mathematically and algorithmically determining how to act intelligently, and presents a burgeoning set of methods that data scientists can apply to industry problems to provide value.

\subsubsection{Challenges}
    
    Game theory and the study of acting intelligently in complex environments with multiple agents is still a highly active area of research today. While still in the early stages of exploration, the field even today offers an interesting solution to the value proposition of how to act intelligently in real world environments. Most of the game theory's advancements come from its application to the study of economics and economics based games. Major challenges still exist as barriers to the widespread application of game theory methods to real world problems. Many of these challenges can be categorized as either complexity based challenges, or modeling based challenges. Computational complexity for finding optimal policies arise due to 
    the exponential or even factorial sizes of game trees, unexpected emergent behavior arising in the game as a whole, constantly changing and non-stationary dynamics, and all of the same difficulties that arose in standard Reinforcement Learning are multiplied when those same algorithms are applied to multi-agent environments. Challenges to formalization arise because games in game theory typically need a fixed set of rules, a fixed set of actions, and a complex multi-variate utility function to perform learning. Real world problems rarely have a fixed set of rules and actions or an easy way to specify the utility of the world. However, even with these major challenges, the value proposition of determining how to act in mathematically optimal ways means this area of study is likely to have ever increasing value for data scientists in industry settings.

\subsection{Example Applications}

\subsubsection{Online and Marketing Applications}

    Reinforcement learning has been applied to many areas of online interactions with customers including search, recommendation, and advertising paper. The paper by Zhao, Xia, Tang, and Yin \cite{Zhao:2019:DRL:3320496.3320500} surveys the increasing role of reinforcement learning for the aforementioned use cases. The authors discuss how reinforcement learning is used for determining the content to return (this is the action) given a search query, the ranking of query results, web page layout optimization, and session based search optimization where one query is temporally and contextually related to a previous query. Much of the work proposes small extensions to reinforcement learning algorithms for the specific application. All are formulated as markov decision processes, where the reinforcement learning problem is slightly modified either in optimization metric, or optimization algorithm that takes into account some of the unique factors of the problem. Reinforcement learning is used in the recommendation case for optimizing recommendations that maximize long term user engagement, show diversity but also remain within a fixed 'theme', and incorporate temporal context. Online advertising challenges lay along a similar vein as the search and content recommendation challenges. Recommending advertisements that are grouped within a theme and most likely to create a purchase is one of the most challenging and most valuable business propositions for the online environment.
    
    In this 2015 paper \cite{Theocharous:2015:PAR:2832415.2832500}, Theocharous, Thomas, and Ghavamzadeh outline an algorithm developed to recommend ads that maximize 'life-time value', or the total number of times a user will click over multiple visits. Their paper outlines a method for computing good policies in a scalable way, and a set of off-policy techniques to check the validity of a policy before it is deployed to a real environment. They implement Q-learning by estimating state action values using a random forest regression model, weighting values using importance sampling, and set Q values using a bounding mechanism. They show that reinforcement learning can be an effective method for recommending ads that optimize long term metrics such as life-time value. These are just a few of the many ways in which reinforcement learning is being applied to online and marketing business problems.

\subsubsection{Customer Experience and Engagement}

    Reinforcement learning is also being increasingly applied to improve engagements with customers for reasons other than marketing or ads. Cisco uses reinforcement learning algorithms in the optimization of its digital engagements with customers. Adobe's marketing methods can also transfer to the case of engagements with customers for the purpose of improving a customer's experience. This paper \cite{Sinha2019SurveysWQ} describes how they use online click stream data along with reinforcement learning to generate a value function, which can be used as surrogate for measuring customer satisfaction. They argue that the proxy based approach to measuring customer satisfaction using reinforcement learning allows companies to do three things that can't be done with standard surveys. The first is to generate ratings by the use of nearly all customers, whereas surveys are typically information gathered from only a few customer. The second is that surveys are not tied to any kind of actual customer behavior. If a customer interacts with lots of online content or uses their device a lot more, this specific behavior cannot be tied back to their responses on a survey. The third is that reinforcement learning can show the value of particular sequences of click actions, while surveys naturally roll up all previous customer sentiment into one point. The interesting value proposition from this paper is that reinforcement learning is used as a tool for understanding user behavior and customer experiences, going beyond just the use of reinforcement learning as an automated method for taking actions.
    
    Another paper \cite{DBLP:journals/corr/abs-1805-06254} discusses the case of customer experience management from the use case of digital programmable networks. They discuss how toady's adaptive programmable networks operate through the use of traditional rule-based decision making, and how this may lead to less than optimal customer experiences on the network. The authors do a good job of positioning both the technical and the business use case for developing algorithms that can automatically prescribe actions and make decisions that will increase metrics such as 'Quality-of-Service' and 'Quality-of-Experience'. The authors introduce the idea of using an iterative learning algorithm like reinforcement learning to attempt to solve the high level challenges with automatically making technical decisions that are likely to improve customers experiences when interacting with a network. 

    A third paper \cite{DBLP:journals/corr/abs-1905-02219} describes how Microsoft used contextual bandits in an attempt to implement a customer support chat bot for goal directed dialogue. They utilized reinforcement learning to handle the diversity of customer intents that exist, the variety of Microsoft products, and the combination of both customer and business value metrics such as solving a customer problem and increasing content click rate. Their work borrows from other applications of reinforcement learning in the wild, and brings up some important points about reinforcement learning systems that need to learn to interact in an environment where no simulation exists initially. Some of their guiding insights include starting with simplified action spaces, starting with imitation learning, intelligent exploration, support for environment changes, and a focus on reward shaping to name a few.

    The customer experience and engagement case uncovers a major challenge of applying reinforcement learning. It is challenging to discover the best algorithm, parameters, and hyper-parameters to use for a specific engagement case. It is common practice to use simulation environments to tune and customize the parameters for nearly all reinforcement learning methods. However a simulation environment typically doesn't exist out of the box. A way around this may be to attempt to create a simulated environment, but that can be challenging because we often already assume the model of the environment is too challenging to model, which is why we used the model free methods of RL in the first place. In application, modern reinforcement learning creates a chicken and egg problem. We need to pick an algorithm, parameters, and hyper-parameters to apply reinforcement learning to an environment. However, we must test and check our choices to determine the right combination for a particular environment which leads to excessive testing. This can actually preclude certain methods like deep reinforcement learning when a simulated environment doesn't exist to modify learning algorithms.

\section{Case Study}

    In this section, we outline a common problem encountered by businesses, and how to apply our methodology to maximize the value returned from this data science project. Assume that a given company provides a service to customers for which they must pay a monthly subscription. A well known question in this domain is whether or not a particular customer or group of customers will renew their subscription or not.  This is typically referred to as predicting churn, or churn risk prediction. It is closely related to sales and demand forecasting mentioned in the prediction and pattern mining section. Many data science projects focus simply on answering this question. We show below that value can be gained beyond simply constructing a model that predicts the churn risk of a customer by framing the problem in terms of the big three questions.

\subsection{Question 1: What is happening or will happen?}

    As we've already stated, the main question that is typically posed to a team of data scientists is 'Can we accurately predict which customers will renew and which ones won't?' While this is a primary question asked by the business, there are many other questions that fall into the area of prediction and pattern mining including,

    \begin{enumerate}
	    \item How much revenue can we expect from renewals? What does the distribution look like?
	    \item What's the upper/lower bound on the expected revenue predicted by the models?
	    \item What are the similar attributes among customers likely to churn versus not churn?
	    \item What are the descriptive statistics for customers likely to churn vs not churn collectively, in each label grouping, and in each unsupervised grouping?
    \end{enumerate}
    
    Each of the above questions can be answered systematically by framing them as problems either in prediction or pattern mining, and by using the wide variety of mathematical methods found in the referenced materials from that section. These are the questions and methods data scientists are most familiar, and will most commonly be answered for a business.

\subsection{Question 2: Why is this happening or going to happen?}

    Given this first question, the immediate next question is why. Why are customers likely or not likely to churn? For each question that we can build a model for, we can also perform a causal analysis. Thus, we can already potentially double the value that a data science project returns by simply adding on a causal analysis to each predictive model built. It's important to bring up again that this question is so important that most data scientists are either answering it incorrectly, or are misrepresenting the information from statistical associations. 
    
    Specifically, when a data scientist is asked the question of why a customer is likely to churn, they almost exclusively turn to feature importance and local models such as LIME \cite{lime} and others. These methods for describing the reason for a prediction are almost always incorrect because there is a disconnect between what the business stakeholder is asking for and what the data scientist is providing because of two different interpretations of the term 'why'. Technically, one can argue that feature importance measures what features are important to 'why' a model makes a prediction, and this would be correct. However, a business stakeholder is nearly always asking the question of 'what is causing the metric itself' and not 'what is causing the metric prediction'. The business stakeholder wants to know the causal mechanisms for why a metric is a particular number. This is something that feature importance absolutely does not answer. The stakeholder wants to use the understanding of the causal mechanisms to take an action to change the prediction to be more in their favor. This requires a causal analysis. However, most data scientists simply take the features with highest measured importance and present them to the stakeholder as though they are the answer to their causal question. This is objectively wrong, yet is time and again presented to stakeholders by seasoned statisticians and data scientists. 
    
    The issue is compounded by the further confusion added by discussions around 'interpretable models' and by the descriptions of feature importance analysis. LIME goes so far as to say that is 'explaining what machine learning classifiers (or models) are doing'. While still a technically correct statement, these methods are being used to incorrectly answer causal questions, leading stakeholders to take actions that can often have the opposite effect of what they intended.
    
    While we've outlined the main causal question, there are a number of questions that can also be asked, and corresponding analysis that can be performed including, 
    
    \begin{enumerate}
        \item How are variables correlated with each other and the churn label? (A non-causal question)
    	\item What are the important features for prediction in a model in general? (A non-causal question)
    	\item What are the most important features for prediction for an individual? Do groupings of customers with locally similar relationships exist? (A non-causal question)
    	\item What are the possible confounding variables? (A causal question)
    	\item After controlling for confounding variables, how do the predictions change? (A non-causal question benefiting from causal methods)
    	\item What does the causal bayes net structure look like? What are all of the reasonable structures? (A causal question)
    	\item What are the causal effect estimates between variables? What about between variables and the class label? (A causal question).
    \end{enumerate}
    
    Many of these questions can be answered in whole or in part by a thorough causal analysis using the methods we outlined in the corresponding causal inference section, and further multiply the value returned by a particular data science project. 

\subsection{Question 3: How can we take action to make the metrics improve?}

    The third question to answer is 'what actions can a stakeholder take to prevent churn?' This is ultimately the most valueable of the three questions. The first two question set the context for who to focus on and where to focus efforts. Answering this question provides stakeholders with a directed and statistically valid means to improve the metrics they care about given complex environments. While still challenging given the methods available today (and presented in the section on intelligent action), it provides one of the greatest value opportunities. Some other questions that can be answered related to intelligent action that stakeholders may be interested in include,

    \begin{enumerate}
    	\item What variables are likely to reduce churn risk if our actions could influence them?
	    \item What actions have the strongest impact on the variables that are likely to influence churn risk, or to reduce churn risk directly?
	    \item What are the important pieces of contextual information relevant for taking an action?
	    \item What are the new actions that should be developed and tested in an attempt to influence churn risk?
	    \item What actions are counter-productive or negatively impact churn risk?
	    \item What does the diminishing marginal utility of an action look like? At what point should an action no longer be taken?
	\end{enumerate}

    The right method to use for prescribing intelligent action depends largely on the problem and the environment. If the environment is complex, the risks high, and there is not much chance for an automated system to be implemented, then something like decision theory, influence diagrams, and game theory based analysis are good candidates. However, if a problem and stakeholder are open to the use of an automated agent to learn and prescribe intelligent actions, then reinforcement learning may be a good choice. While possibly the most valuable of the big three questions to answer, it also exists as one of the most challenging. There still many open research questions related to answering this question, but the value proposition means that it's likely an area that will see increased industry investment in the coming years.

\section{Conclusions}

    The value of answering causal and intelligent action questions are real. They are so important that current predictive and statistical methodologies are miss-applied in an attempt to answer them. They are the natural next step after any supervised or unsupervised prediction or pattern mining task. The work that has been done in the fields of causal inference, decision theory, and reinforcement learning will begin surface as a part of every project in business applications. It's likely that in the future, every data science modeling or analysis project will have 3 distinct phases, the first being to build a classification or regression model, the second being to perform causal inference, and the third being to build decision networks, reinforcement learning algorithms, or some other intelligent action system to improve desired metrics. 
    
    The barriers to achieving this inevitable result include a lack of knowledge and teaching about the methodologies, and possibly limitations with the technologies and methods themselves. Causal inference requires an uncommon statistical understanding, decision theory and influence diagrams require formalizing a problem's structure, and reinforcement learning is currently sensitive to the problem formulation and can be data intensive. However, regardless of the specific methods that are available today, the need to understand and answer 'The Big Three' questions in mathematically optimal ways will remain. The value that Data Science, Machine Learning, and AI can bring in the coming years will be in developing methods to provide answers to these value oriented questions, as well as in the automation of the processes to produce these answers.

\bibliographystyle{plain} % We choose the "plain" reference style
\bibliography{refs} % Entries are in the "refs.bib" file

% Note, for submission to axriv, use 'submit' to download zip, then 
% replace \bibliography{refs} with the text from the main.bbl file.

\end{document}